\documentclass[letterpaper, 10 pt, journal, twoside]{IEEEtran}
\usepackage{caption}
\usepackage{booktabs}
\usepackage{pdfpages}
\usepackage{epstopdf}
\usepackage{graphicx}%
\usepackage{multirow}%
\usepackage{amsmath,amssymb,amsfonts}%
\usepackage{amsthm}%
\usepackage{mathrsfs}%
\usepackage{bm}
\usepackage{xcolor}%
\usepackage{textcomp}%
\usepackage{manyfoot}%
\usepackage{booktabs}%
\usepackage{algorithm}%
\usepackage{algorithmicx}%
\usepackage{algpseudocode}%
\usepackage{listings}%
\usepackage{soul}
\usepackage{subcaption}
\usepackage{url}
\usepackage{flushend}
\usepackage{hyperref}
\usepackage{cases}
\usepackage{pifont}% http://ctan.org/pkg/pifont
\newcommand{\cmark}{\ding{51}}%
\newcommand{\xmark}{\ding{55}}%

\makeatletter
\newcommand{\beginsupplement}{
  \setcounter{section}{0}
  \renewcommand{\thesection}{S\arabic{section}}

  \renewcommand{\thesubsection}{\thesection.\arabic{subsection}}
  \renewcommand{\subsection}{\@startsection{subsection}{2}{\z@}%
    {0.5\baselineskip}{0.5\baselineskip}{\normalfont\normalsize\itshape}}
  \renewcommand{\thesectiondis}{\thesection} % Override IEEE's internal formatting
  \renewcommand{\thesubsectiondis}{\thesubsection} % Force IEEE to use numeric subsections

  \setcounter{subsection}{0} % Reset subsection counter
  
  \setcounter{equation}{0}
  \renewcommand{\theequation}{S\arabic{equation}}
  \setcounter{table}{0}
  \renewcommand{\thetable}{S\arabic{table}}
  \newcounter{offset}
  \setcounter{offset}{\value{figure}}
  \renewcommand{\thefigure}{S\the\numexpr\value{figure}-\value{offset}\relax}
}
\makeatother

\begin{document}

\title{
Learning to Adapt through Bio-Inspired Gait Strategies for Versatile Quadruped Locomotion
}

\author{Joseph Humphreys$^{1}$ and Chengxu Zhou$^{2}$
\thanks{This work was partially supported by the Royal Society [grant number RG\textbackslash R2\textbackslash232409] and the Advanced Research and Innovation Agency [grant number SMRB-SE01-P06].
\textit{(Corresponding author: Chengxu Zhou)}}
\thanks{$^{1}$School of Mechanical Engineering, University of Leeds, UK, LS2 9JT. {\tt\footnotesize el20jeh@leeds.ac.uk}}% <-this % stops a space
\thanks{$^{2}$Department of Computer Science, University College London, UK, WC1E 6BT. {\tt\footnotesize chengxu.zhou@ucl.ac.uk}}% <-this % stops a space
}
\maketitle

\begin{abstract}
Legged robots must adapt their gait to navigate unpredictable environments, a challenge that animals master with ease. However, most deep reinforcement learning (DRL) approaches to quadruped locomotion rely on a fixed gait, limiting adaptability to changes in terrain and dynamic state. Here we show that integrating three core principles of animal locomotion-gait transition strategies, gait memory and real-time motion adjustments enables a DRL control framework to fluidly switch among multiple gaits and recover from instability, all without external sensing. Our framework is guided by biomechanics-inspired metrics that capture efficiency, stability and system limits, which are unified to inform optimal gait selection. The resulting framework achieves blind zero-shot deployment across diverse, real-world terrains and substantially significantly outperforms baseline controllers. By embedding biological principles into data-driven control, this work marks a step towards robust, efficient and versatile robotic locomotion, highlighting how animal motor intelligence can shape the next generation of adaptive machines.
\end{abstract}


%%%%%%%%%%%%%%%%%%%%%%%%%%%%%%%%%%%%%%%%%%%%%%%%%%%%%%%%%%%%%%%%%%%%%%%%%%%%%%%%%%%%%%%%%

Originally inspired by the remarkable adaptability of quadruped mammal locomotion, an ability shaped by innate and environmentally induced factors \cite{innateGait, innateGait2, gaitEvo}, the field of quadruped robotics has invested substantial resources into developing equally proficient locomotion frameworks. At present, the most advanced systems rely on end-to-end deep reinforcement learning (DRL), which involves training a multilayer perceptron \cite{mlp} capable of navigating diverse environments. 

These frameworks demonstrate impressive realisation of terrestrial locomotion skills, which can be classified into two groups (further detailed in Supplementary Section 1); Froude-characterised locomotion (no desired velocity normal to the ground plane and hence upholds the assumptions of the Froude number \cite{Froude2}) such as walking or running in real-world \cite{quadWild, terrdeform}, urban \cite{RLroughterrain, RLfallrecovery} and deformable\cite{terrdeform} terrains, and Froude-free locomotion (features a desired velocity normal to the ground plane) such as jumping between platforms \cite{quadjump}, climbing over obstacles \cite{parkour}, and sure-footedness \cite{deeptrackingcontrol}. Despite these achievements, when it comes to Froude-characterised locomotion, which can account for about $\approx70-90\%$ daily animal locomotion \cite{locopercent1, locopercent2}, the adaptability of these frameworks remains constrained, as most systems are limited to deploying a single targeted gait or locomotion strategy.

\begin{figure*}
\centering
\includegraphics[width=1.0\textwidth]{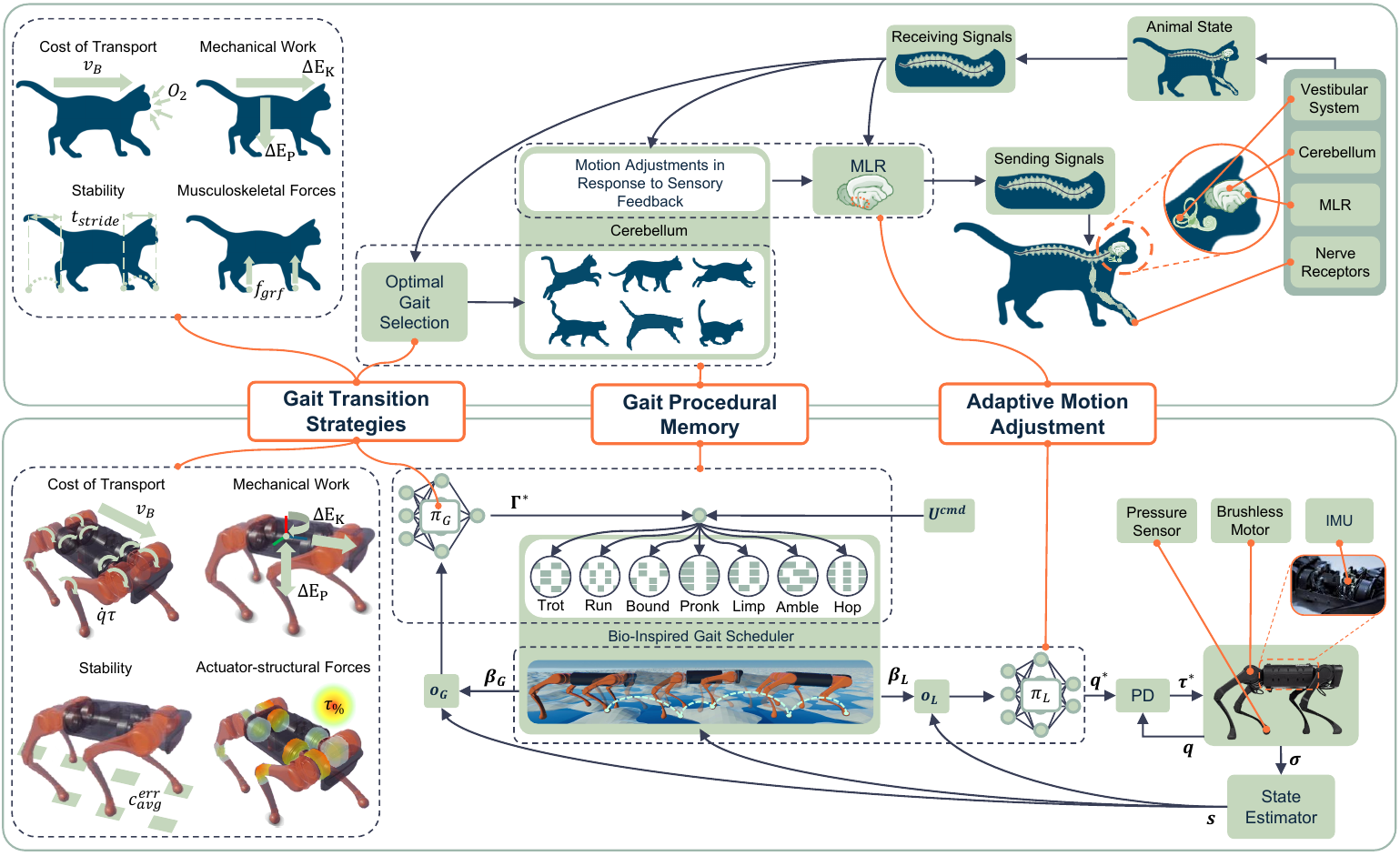}
\vspace{-2mm}
\caption{\textbf{Instilling the core animal locomotion proficiency attributes within a DRL locomotion framework.} From taking an abstracted view of animal locomotion to determine the attributes of proficient locomotion, we instil animal gait transition strategies within the gait selection policy, $\pi_{G}$, gait procedural memory is embedded within the bio-inspired gait scheduler, and adaptive motion adjustments are realised by the locomotion policy, $\pi_{L}$. $\pi_{G}$ has been trained to minimise the animal gait transition metrics applied to the quadruped robot based on the current robot state, $\bm{s}$, and relevant bio-inspired gait scheduler output, $\bm{\beta}_{G}$, to select the optimal gait, $\Gamma^{*}$. The bio-inspired gait scheduler then generates the gait references informed by $\bm{s}$ from encoded high-level gait parameters. The gait references, $\bm{\beta}_{L}$, are then passed to $\pi_{L}$ to inform of any adjustments to the nominal gait motions. The control framework is constant between training and deployment to improve sim-to-real transfer. Regarding acronyms, MLR is, mesencephalic locomotor region, PD is
proportional derivative controller, and IMU is inertial measurement unit.}
\vspace{-2mm}
\label{fig:concept}
\end{figure*}

%\captionof{figure}{\textbf{Instilling the core animal locomotion proficiency attributes within a DRL locomotion framework.} From taking an abstracted view of animal locomotion to determine the attributes of proficient locomotion, we instil animal gait transition strategies within the gait selection policy, $\pi_{G}$, gait procedural memory is embedded within the bio-inspired gait scheduler, and adaptive motion adjustments are realised by the locomotion policy, $\pi_{L}$. $\pi_{G}$ has been trained to minimse the animal gait transition metrics applied to the quadruped robot based on the current robot state, $\bm{s}$, and relevant bio-inspired gait scheduler output, $\bm{\beta}_{G}$, to select the optimal gait, $\Gamma^{*}$. The bio-inspired gait scheduler then generates the gait references informed by $\bm{s}$ from encoded high-level gait parameters. The gait references, $\bm{\beta}_{L}$, are then passed to $\pi_{L}$ to inform of any adjustments to the nominal gait motions. The control framework is constant between training and deployment to improve sim-to-real transfer.}
%\label{fig:concept}

In contrast, biomechanics research has shown that no single gait is universally optimal across all scenarios within Froude-characterised locomotion \cite{wildebeestTerrainGait, cheetaTerrainGait, dogTerrainGait}. Animals adapt their locomotion by employing nominal gaits such as ambling, trotting, and running \cite{animalgaits}, while switching to specialised gaits such as hopping, pronking and bounding for off-nominal tasks such as predator evasion or obstacle navigation \cite{pronkbound}. Current DRL frameworks fall short of replicating this level of Froude-characterised locomotion versatility. To address this limitation, some approaches have focused on training DRL policies to learn multiple gaits by providing reference motions during training \cite{RLgaitmodular, RLmpcref, RLe2eGaitTrans, RLmanyrefs}, or by learning from policies that specialise in specific gaits \cite{RLenergymin}. However, these methods remain insufficient when compared to the extensive capabilities observed in animal locomotion, which include:
\begin{itemize} 
    \item Adaptation of gait style for optimal performance in response to challenging terrains and perturbations. \begin{itemize} 
        \item Enabled by advanced gait selection strategies. 
    \end{itemize} 
    \item Rapid deployment of a diverse set of task- and state-specific gaits. 
    \begin{itemize} 
        \item Attributable to gait procedural memory. 
    \end{itemize} 
    \item Seamless deviation from nominal gait motions to address off-nominal contact states. 
    \begin{itemize} 
        \item Achieved through precise motion adjustments tailored to the environment. 
    \end{itemize} 
\end{itemize}
Although existing DRL frameworks have demonstrated progress in implementing learned gaits, none successfully integrate all three attributes simultaneously. This gap highlights the considerable potential of biomechanics-inspired approaches to advance robotic locomotion.

While bio-inspired methods that leverage central pattern generators (CPGs) \cite{CPG} have realised spontaneous gait transitions \cite{CPG2} and mimic certain animal behaviours \cite{natureCommunications}, their performance in real-world applications is often limited. Typically, such experiments are constrained to controlled environments \cite{CPG3, natureCommunications} or, when conducted outdoors, are restricted by low velocities and simple tasks \cite{RLe2eGaitTrans}. Another bio-inspired approach is training locomotion policies from re-targeted animal motion data \cite{lifelike}. Although this method can achieve natural Froude-free locomotion, there is no guarantee that the exhibited Froude-characterised locomotion is optimal; it is unreasonable to assume that mechanically different systems would perform efficiently with the same low-level behaviour. These limitations suggest that instead of attempting to replicate animal locomotion mechanisms precisely, state-of-the-art DRL frameworks should be augmented with high-level attributes derived from animal locomotion to instil the proficiency observed in nature.

Animal gait transition strategies, which contribute to optimal performance and enable navigation of challenging environments, are believed to emerge from the minimisation of metrics related to energy consumption \cite{CoTTrans, CoTTrans2, Froude2}, mechanical work \cite{mechwork, mechwork2, mechwork3}, instability \cite{stabTrans, mouseGait}, and musculoskeletal forces \cite{mechanicaltrigger, mechanicaltrigger3, mechanicaltrigger2}. However, no singular metric has been definitively identified as the sole driver of these transitions. Instead, it is hypothesised that a combination of these factors influences gait transition strategies \cite{stabTrans, notrigger1, notrigger2}.

The concept of gait procedural memory, which facilitates the rapid deployment of a range of gaits, is thought to reside within the cerebellum of the animal brain. This region governs the coordination of limb movements for each gait learned by the animal \cite{locobraindecoding, animalgaitcontrol}. Similarly, adaptive motion adjustments, crucial for seamless adaptation to off-nominal contact states, are achieved through coordination between the mesencephalic locomotor region (MLR), which oversees locomotion execution \cite{mlr}, and the cerebellum. These adjustments rely on sensory feedback to modify limb movements in response to the animal's current state \cite{cerebellarcontrol}.

Despite these insights, there has been no prior attempt to simultaneously integrate all these attributes---gait transition strategies, procedural memory, and adaptive motion adjustments---within existing DRL locomotion frameworks. This leads to the following research questions:
\begin{enumerate} 
    \item How can the roles of the MLR and cerebellum inspire the augmentation of an end-to-end DRL locomotion policy to adapt to off-nominal contact states?
    \item Can a DRL locomotion policy, inspired by gait procedural memory, learn to deploy a diverse set of gaits and perform rapid gait transitions?
    \item How can metrics that characterise animal gait transition strategies be effectively leveraged within a DRL policy for optimal gait selection? Does the resulting behaviour align with that observed in animals?
    \item Can the developed framework exhibit exemplary adaptability to traverse real-world terrains not encountered during training? What is the contribution of each metric to this adaptability?
\end{enumerate}

\begin{figure*}[h]
    \centering
    \includegraphics[width=1\textwidth]{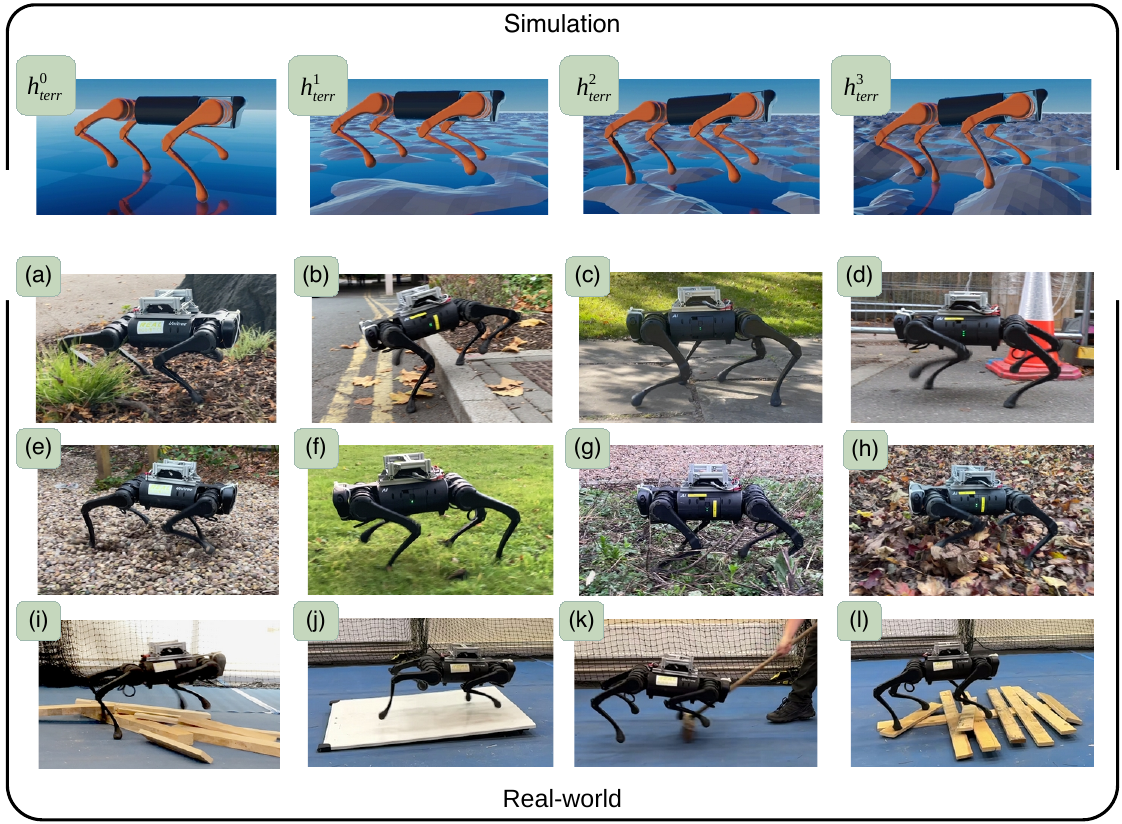}
    \centering
    \vspace{-5mm}
    \caption{\textbf{Snapshots of our framework being deployed in different environments.} Simulated terrains $h_{\text{terr}}^{0}$ through $h_{\text{terr}}^{3}$ are generated with fractal noise with maximum heights of $0$ m, $0.06$ m, $0.13$ m and $0.2$ m. To demonstrate the adaptability of our framework is transferable to the real world it has been deployed on (a) wood-chip, (b) a large step, (c) concrete slabs with large cracks, (d) tarmac, (e) deep rocks, (f) grassy terrain, (g) overgrown roots, (h) fallen leaves, (i) loose timber, (j) low-friction ramp, (k) flat terrain with perturbations, and (i) balanced timber.} 
    \label{fig:terr_sc}
\end{figure*}

To address these questions, we propose a novel DRL locomotion framework (see Figure \ref{fig:concept}) designed to incorporate these key animal locomotion attributes. The framework demonstrates exceptional adaptability through zero-shot deployment in complex, real-world environments, relying solely on interoceptive sensors.

%%%%%%%%%%%%%%%%%%%%%%%%%%%%%%%%%%%%%%%%%%%%%%%%%%%%%%%%%%%%%%%%%%%%%%%%%%%%%%%%%%%%%%%%%
\section{Results}
Within the framework, presented in Figure \ref{fig:concept}, a gait selection policy, $\pi_{G}$, is trained for optimal gait selection through minimising gait transition metrics adopted from biomechanics to generate the output $\Gamma^{*} \in [0, 7]$ which maps to a specific gait within $[\textit{stand},\textit{trot},\textit{run},\textit{bound},\textit{pronk},\textit{limp},\textit{amble},\textit{hop}]$. This selected gait, coupled with the velocity command within $\bm{U}^{\text{cmd}}=[v_{x}^{\text{cmd}},v_{y}^{\text{cmd}},\omega_{z}^{\text{cmd}},\Gamma^{*}] \in {\mathbb{R}}^{\text{4}}$, where $v_{x}^{\text{cmd}}$, $v_{y}^{\text{cmd}}$ and $\omega_{z}^{\text{cmd}}$ are base velocities in $x$, $y$ and yaw respectively, is passed to the bio-inspired gait scheduler (BGS) to generate gait references (along with transition references between any gait pair), as detailed in Section \ref{subsec:BGS}. These gait references are contained within the BGS outputs $\bm{\beta}_{L}$ and $\bm{\beta}_{G}$ for the locomotion policy, $\pi_{L}$, and $\pi_{G}$ respectively through inclusion within their observation vectors $\bm{o}_{L}$ and $\bm{o}_{G}$. In this respect, the BGS acts as pseudo gait procedural memory. The gait references are adjusted based on the robot's state, $\bm{s}$, generated by the state estimator (SE) from the sensor feedback vector, $\bm{\sigma}$, which in turn reflects the relationship between the cerebellum and MLR. To realise the output joint positions of $\pi_{L}$, $\bm{q}^{*}$, they are passed through a proportional derivative controller to generate joint torque commands $\bm{\tau}^{*}$.

We evaluate the proposed framework through a set of studies, demonstrating it outperforms others by exhibiting quadruped animal locomotion strategies, and validating this gained proficiency on real-world terrain, as presented in Figure \ref{fig:terr_sc} and Supplementary Video S1.

\subsection{Achieving Adaptive Motion Adjustment with a Diverse Set of Gaits}
\label{subsec:loco_bio_exp}

\begin{figure*}
    \centering
    \includegraphics[width=0.9\textwidth]{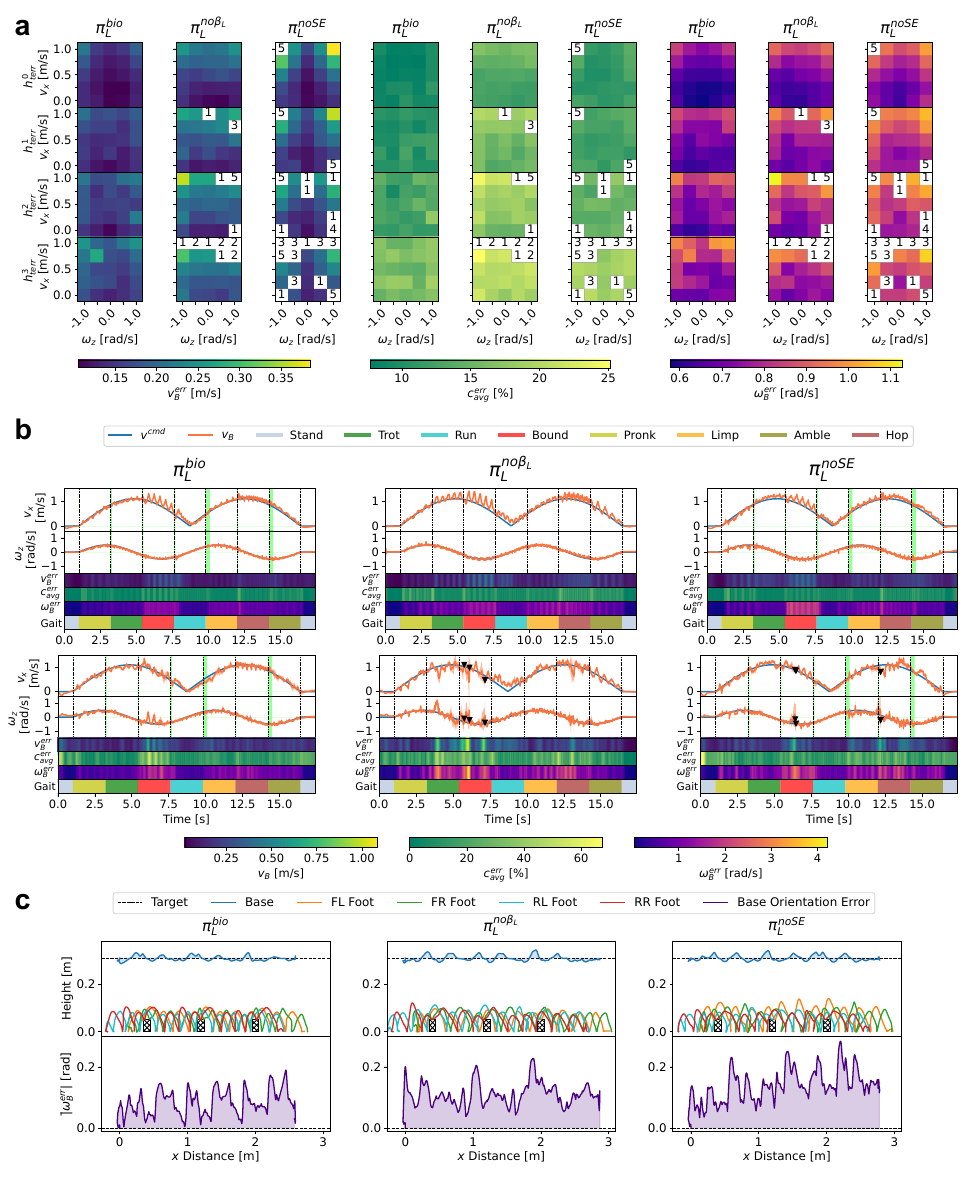}
    \centering
    %\vspace{-5mm}
    \caption{\textbf{Locomotion comparison study experiments} (a) each policy follows a set of command velocities in $x$ and yaw between $0$ to $1$ m/s and $-1$ to $1$ rad/s respectively. During each velocity pair, the commanded gait is cycled through all gaits, switching every $1$ s. This repeated $5$ times over flat, $h_{\text{terr}}^{0}$, to very rough terrain, $h_{\text{terr}}^{3}$ (shown in Figure \ref{fig:terr_sc}) with the average performance being plotted. A number rather than a magnitude indicates the count of experiments that the policy failed. (b) each policy follows a sinusoidal trajectory in $x$, $v_{x}$, and in yaw, $\omega_{z}$, while switching gaits every $2$ s, which is repeated $5$ times over $h_{\text{terr}}^{0}$ and $h_{\text{terr}}^{3}$ terrain. Green highlighted areas indicating transition phases, for which the function and formulation are detailed in Section \ref{subsec:BGS}, and black triangles representing points of failure. (c) policies follow a command of just $v_{x}=0.5$ m/s while using a constant trot gait while encountering rectangular steps with a height of $0.05$ m, where $|\bar{\omega}_{B}^{\text{err}}|$ is the magnitude of the desired orientation error. FL is front left, FR is front right, RL is rear left, and RR is rear right in reference to the robot's feet.}
    \label{fig:loco_policy_study}
\end{figure*}

To evaluate our method of instilling adaptive motion adjustment and procedural memory for diverse gait deployment, a comparison study is completed between our bio-inspired locomotion policy $\pi_{L}^{\text{bio}}$, a standard multi-gait locomotion policy with no pseudo procedural memory $\pi_{L}^{\text{no}\bm{\beta}_{L}}$, and a policy trained which also uses $\bm{\beta}_{L}$ within $\bm{o}_{L}$ but implements the standard approach of extracting the observations from the simulator, $\pi_{L}^{\text{noSE}}$, as shown in Supplementary Video S2. From the results of this study, shown in Figure \ref{fig:loco_policy_study}, the proficiency of $\pi_{L}^{\text{bio}}$ over $\pi_{L}^{\text{no}\bm{\beta}_{L}}$ and $\pi_{L}^{\text{noSE}}$ in terms of velocity tracking error $v_{B}^{\text{err}}$, contact schedule tracking error $c_{\text{avg}}^{\text{err}}$, and base stability (magnitude of undesirable base angular velocities) $\omega_{B}^{\text{err}}$, is stark. 

As shown in Figure \ref{fig:loco_policy_study}a, on flat terrain, $\pi_{L}^{\text{bio}}$ achieves lower $v_{B}^{\text{err}}$, $c_{\text{avg}}^{\text{err}}$, and $\omega_{B}^{\text{err}}$ than $\pi_{L}^{\text{no}\bm{\beta}{L}}$ and $\pi_{L}^{\text{noSE}}$, averaging $15\%$, $21\%$, and $10\%$ lower respectively (excluding failures), with errors increasing at higher velocity commands. This performance gap widens on rough terrain, where $\pi_{L}^{\text{no}\bm{\beta}{L}}$ and $\pi_{L}^{\text{noSE}}$ fail frequently, especially at higher speeds and rougher surfaces, while $\pi_{L}^{\text{bio}}$ succeeds in all trials despite being trained only on flat terrain. This highlights its adaptability to unseen environments (see Section \ref{subsec:policytraining} for rationale behind omitting rough terrain in $\pi_{L}$ training).

This incompetence of $\pi_{L}^{\text{no}\bm{\beta}_{L}}$ and $\pi_{L}^{\text{noSE}}$ is caused by a lack of adaptive swing foot motion adjustments captured within $\bm{\beta}_{L}$ based on $\bm{s}$ and the accumulation of error within the SE respectively. Both of these factors are substantially affected by the instabilities rough terrain inflicts upon the robot. Considering that the nominal swing foot peak height is defined as $25\%$ of the nominal base height and for $h_{\text{terr}}^{2}$ and $h_{\text{terr}}^{3}$ (as defined in Figure \ref{fig:terr_sc}) the peak terrain height is $44\%$ and $67\%$ of the base height, having no data or strategy to account for this harsh terrain results in the rapid deterioration of proficiency.

For $\pi_{L}^{\text{no}\bm{\beta}{L}}$, Figure \ref{fig:loco_policy_study}b shows large spikes in $c_{\text{avg}}^{\text{err}}$ and $\omega_{B}^{\text{err}}$ when encountering $h_{\text{terr}}^{3}$, highlighting its inability to adapt swing foot trajectories and overcome an unrefined solution space. Figure \ref{fig:loco_policy_study}c further shows sustained instabilities in base height and orientation after contact with steps at $17\%$  of the nominal base height. For $\pi_{L}^{\text{noSE}}$, Figs. \ref{fig:loco_policy_study}b and \ref{fig:loco_policy_study}c show that poor reference tracking and stability worsen with velocity command magnitude and time, as it lacks strategies to mitigate error build-up in the SE. Since $\pi_{L}^{\text{bio}}$ does not experience any of these limitations, this explicitly demonstrates the effectiveness of implementing $\bm{\beta}_{L}$ and $\bm{s}$ within $\bm{o}_{L}$; $\pi_{L}^{\text{bio}}$ successfully generalises across gaits and terrain, demonstraing successful instillation of adaptive motion adjustment and gait procedural memory.

\begin{figure*}[h]
    \centering
    \includegraphics[width=1\textwidth]{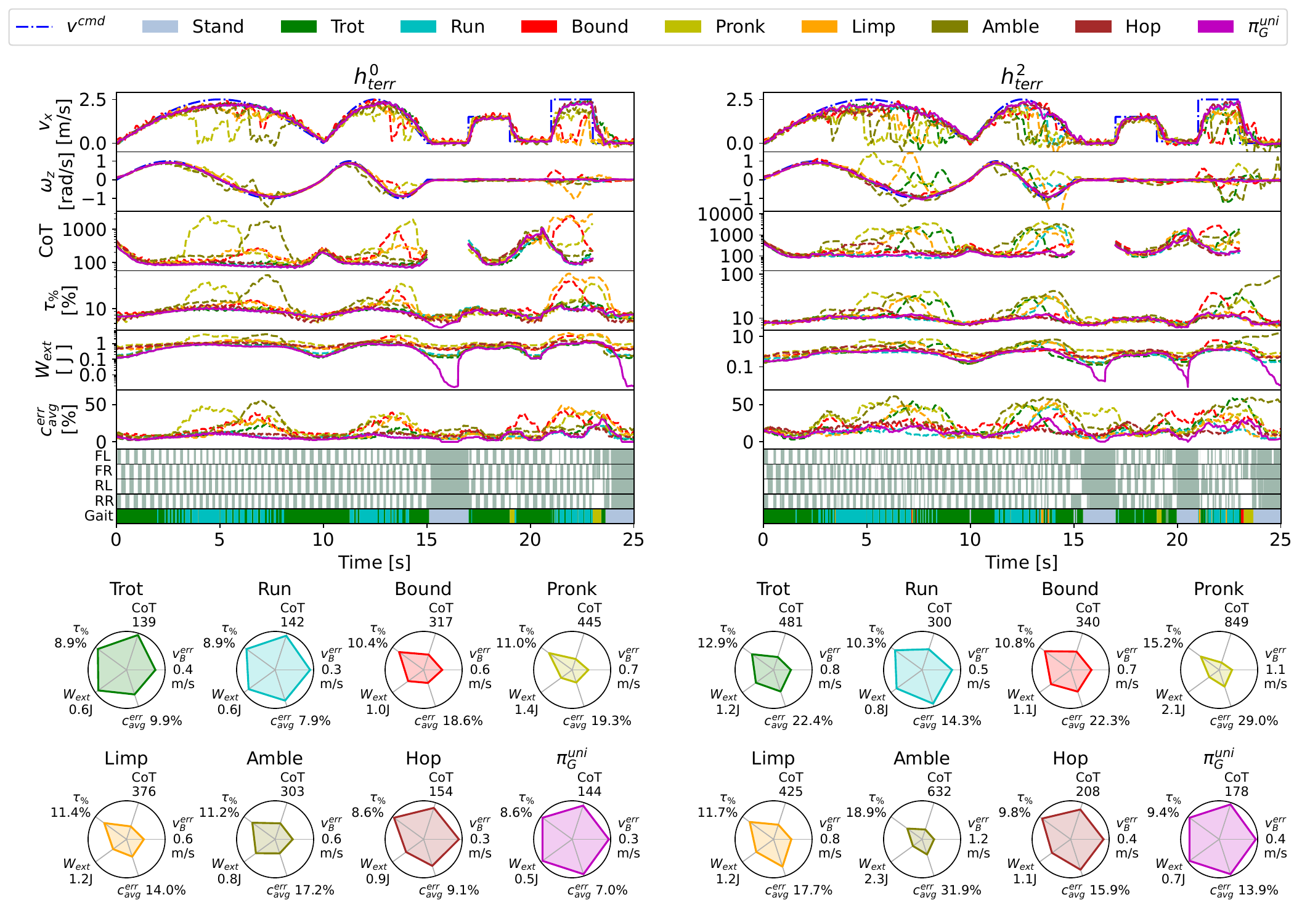}
    \centering
    \vspace{-5mm}
    \caption{\textbf{Comparative study between each gait and} $\pi_{G}^{\text{uni}}$. For terrains $h_{\text{terr}}^{0}$ and $h_{\text{terr}}^{2}$, each isolated gait and $\pi_{G}^{\text{uni}}$ are given a velocity command to follow to assess their performance in terms of CoT, $\tau_{\%}$, $W_{\text{ext}}$, and $c_{\text{avg}}^{\text{err}}$; for each individual gait $\pi_{L}^{\text{bio}}$ just run with the gait statically selected and for $\pi_{G}^{\text{uni}}$ it is coupled with $\pi_{L}^{\text{bio}}$ for autonomous optimal gait selection. Additionally at the bottom of these time series plots, the contact state of the feet and the gait that $\pi_{G}^{\text{uni}}$ realises is also shown. The average relative performance in terms of these metrics is displayed in the radar plots at the bottom of the figure, where each gait's performance is normalised to that of the best performer for each metric; the higher the value within the radar plot, the more effectively that metric has been minimised.}
    \label{fig:gs_policy_study}
\end{figure*}

\subsection{Applying Biomechanics Metrics For Optimal Gait Selection}
Directly applying biomechanics metrics to instil animal gait transition strategies is unsuitable due to differences between animals and robots, as well as $\pi_{G}$ training requirements. Instead, we use cost of transport ($\text{CoT}$), torque saturation ($\tau_{\%}$), external work ($W_{\text{ext}}$), and foot contact tracking error ($c_{\text{avg}}^{\text{err}}$) to minimise energy use, actuator-structural forces, mechanical work, and instability, respectively. Details and justification of these metrics are in Section \ref{subsec:biometrics}.

In accordance with Section \ref{subsec:gaitselectionpolicy}, these metrics are unified within the training of the gait selection policy $\pi_{G}^{\text{uni}}$ to instil the strategies animals use for optimal gait selection to acheive exemplary adaptability. To investigate whether $\pi_{G}^{\text{uni}}$ effectively minimises these metrics through gait selection, the results of competing the highly demanding velocity command trajectory presented in Figure \ref{fig:gs_policy_study} and Supplementary Video S3 for $\pi_{G}^{\text{uni}}$ paired with $\pi_{L}^{\text{bio}}$ are collected, along with that for all individual gaits deployable by $\pi_{L}^{\text{bio}}$, for flat terrain and terrain $h_{\text{terr}}^{2}$.

Figure \ref{fig:gs_policy_study} shows that $\pi_{G}^{\text{uni}}$ exclusively uses trotting at low speeds and running at high speeds. When accelerating, $\pi_{G}^{\text{uni}}$ oscillates between trotting and running, with an increasing bias towards running, to increase the stride frequency as visualised by the foot contact data in Figure \ref{fig:gs_policy_study}. Consequently, $\pi_{G}^{\text{uni}}$ not only reliably tracks the optimal gait but outperforms individual gaits, which is only aided by its transition to standing during $v_{x}^{\text{cmd}}=0$ $\omega_{z}^{\text{cmd}}=0$ events for minimal $\tau_{\%}$, $W_{\text{ext}}$ and $c_{\text{avg}}^{\text{err}}$. This behaviour, although never targeted, is reflected in animal locomotion strategies, where gait stride frequency increases with speed, and transitional phases blend gaits to minimise energy use \cite{mechwork2}. Under rough terrain and high acceleration, $\pi_{G}^{\text{uni}}$ employs additional gaits to manage instability. This leads to a gait classification: trot and run serve as nominal gaits for low and high speeds, while bound, pronk, limp, amble, and hop act as auxiliary gaits for off-nominal conditions like stability recovery.

When inspecting the radar charts in Figure \ref{fig:gs_policy_study}, which depict relative performance in terms of the metrics, the origin of this emerged gait selection strategy becomes clear. Across the gaits, on flat terrain trot and run gaits achieve the best relative performance. However, when it comes to overcoming rough terrain bound, hop and limp gaits all gain relative performance in $\tau_{\%}$, $W_{\text{ext}}$ and $c_{\text{avg}}^{\text{err}}$, while trot and run gaits exhibit a reduced dominance in relative proficiency. In addition to this observation providing an insight as to why $\pi_{G}^{\text{uni}}$ chooses to utilise these auxiliary gaits it also provides evidence to suggest that $\tau_{\%}$, $W_{\text{ext}}$ and $c_{\text{avg}}^{\text{err}}$ can effectively characterise stability. This observation is further investigated and discussed in the following sections. Overall, $\pi_{G}^{\text{uni}}$ outperforms all individual gaits across all metrics, with the exception of CoT for trot and run gaits where the difference is negligible, demonstrating the successful minimisation of the metrics and successful instillation of gait procedural memory of how to utilise each gait given the robot's state and task. 

\subsection{Comparison Between Robot and Animal Gait Selection}

\begin{figure*}
    \centering
    \includegraphics[width=0.95\textwidth]{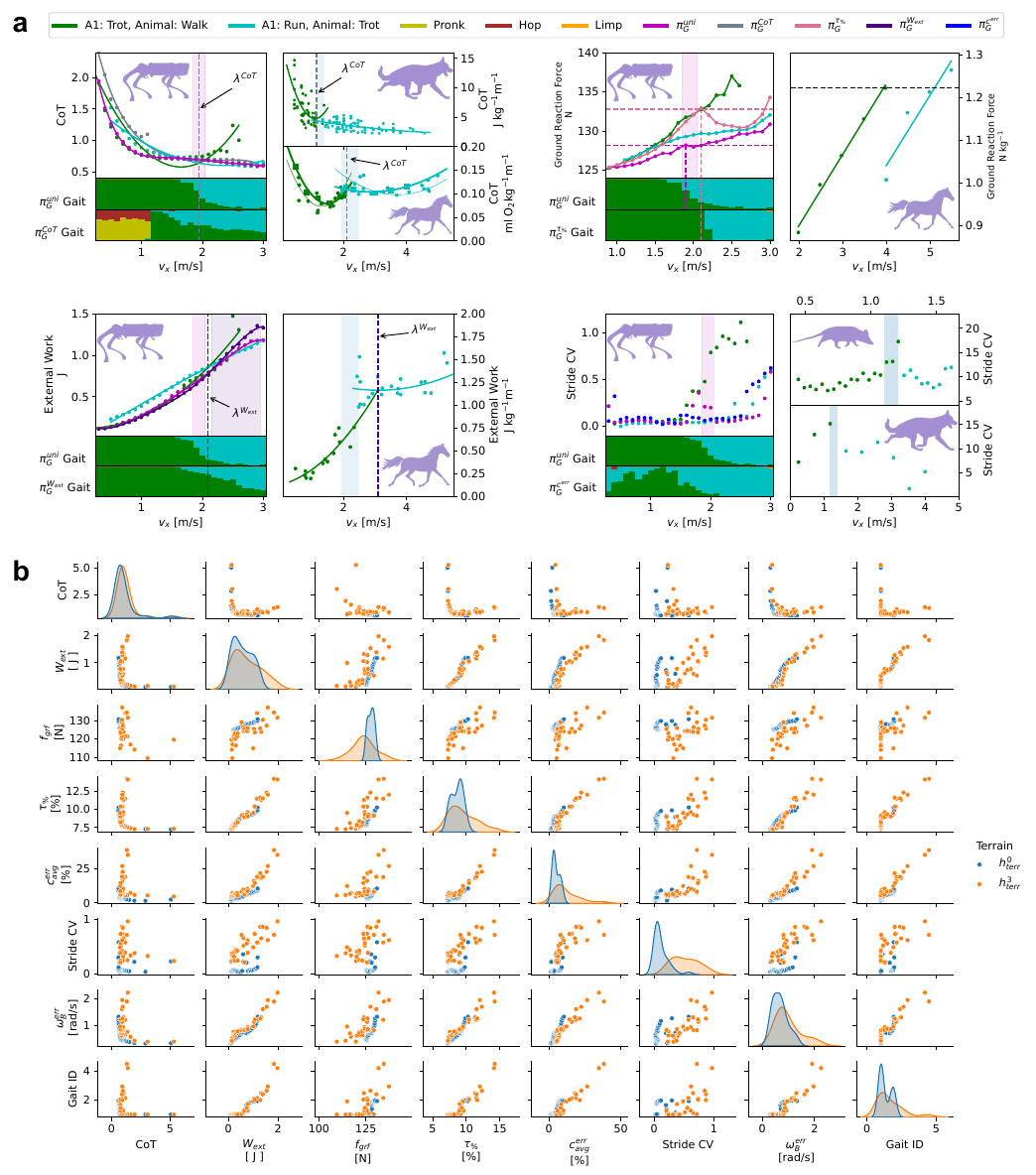}
    \centering
    \caption{\textbf{Comparison between animal and robot gait selection policy strategies} during increasing linear velocity. (a) The bottom two plots of all robot data indicate the percentage of each gait utilised at that velocity. Magenta, purple and blue shaded regions indicate the transition phases of $\pi_{G}^{\text{uni}}$, $\pi_{G}^{W_{\text{ext}}}$ and animals respectively. This study compares transition strategies of (top left) $\pi_{G}^{\text{uni}}$ and $\pi_{G}^{\text{CoT}}$ to data collected from dogs \cite{stabTrans} and horses \cite{Froude1, mechwork2} in terms of CoT, (top right) $\pi_{G}^{\text{uni}}$ and $\pi_{G}^{\tau_{\%}}$ to data collected from horses \cite{mechanicaltrigger} in terms of foot ground reaction forces, (bottom left) $\pi_{G}^{\text{uni}}$ and $\pi_{G}^{W_{\text{ext}}}$ to horses \cite{mechwork2} in terms of external work, and (bottom right) $\pi_{G}^{\text{uni}}$ and $\pi_{G}^{c^{\text{err}}}$ to opossums and dogs \cite{stabTrans}. (b) Mapping the correlation between metrics and the average gait ID selected across velocities for terrains $h_{\text{terr}}^{0}$ and $h_{\text{terr}}^{3}$, with the dataset of each terrain consisting of $31$ data points averaged over $1000$ samples.}
    \label{fig:metric_study}
\end{figure*}
When developing metrics to characterise gait transitions in animals, data is collected over intervals of increasing forward velocity on flat terrain \cite{Froude1, mechanicaltrigger, mechwork2, stabTrans}. Hence, within Figure \ref{fig:metric_study}a we took the same approach. We also repeat this experiment with $h_{\text{terr}}^{3}$ to investigate correlations between the metrics and the effects of introducing rough terrain, as presented in Figure \ref{fig:metric_study}b.
%To provide further insight into the effects of training $\pi_{G}^{\text{uni}}$ with all metrics rather than just an individual metric, we train four additional $\pi_{G}$ policies that individually minimise energy consumption $\pi_{G}^{\text{CoT}}$, actuator-structural forces $\pi_{G}^{\tau_{\%}}$, mechanical work $\pi_{G}^{W_{\text{ext}}}$ and stability $\pi_{G}^{c^{\text{err}}}$ in accordance with Section \ref{subsec:biometrics}. 
Additionally, we train four further $\pi_{G}$ policies that individually minimise energy consumption $\pi_{G}^{\text{CoT}}$, actuator-structural forces $\pi_{G}^{\tau_{\%}}$, mechanical work $\pi_{G}^{W_{\text{ext}}}$ and stability $\pi_{G}^{c^{\text{err}}}$, in accordance with Section \ref{subsec:biometrics}, to compare their performance with $\pi_{G}^{\text{uni}}$.
%\hl{[CZ] no 'further insight' provided after introducing those individual policies. -OK, I see them in the following subsections now. but to address the first impression doubts, can we say "we train individual policies ... and then compared their performance with $\pi_G$?"}. 
One unanimous observation across Figure \ref{fig:metric_study}a is that animals experience a gait transition phase over a range of velocities \cite{stabTrans, Froude1, CoTTrans3}. This behaviour is reflected in $\pi_{G}^{\text{uni}}$, where we class a transition phase as where no individual gait occupies more than 75\% of the gaits used at a specific speed.

\subsubsection{Energy Expenditure -- Cost of Transport}
An animal's gait transition aligns with the CoT-optimal point, $\lambda^{\text{CoT}}$, to minimise energy cost \cite{Froude2, mechwork2}, as shown in Figure \ref{fig:metric_study}a. $\pi_{G}^{\text{uni}}$ mirrors this, tracking the lowest-CoT gait with transitions centred on $\lambda^{\text{CoT}}$. In contrast, $\pi_{G}^{\text{CoT}}$ lacks a defined transition phase and switches earlier, due to training on rough terrain where hopping improves CoT (Figure \ref{fig:gs_policy_study}). Figure \ref{fig:metric_study}b further shows $h_{\text{terr}}^{3}$ induces similar CoT distributions but with greater gait variance, as auxiliary gaits become more effective. This highlights that CoT-only training reduces generality and diverges from natural gait selection.

\subsubsection{Actuator-structural Forces -- Foot Contact Forces}
% Reason why $\pi_{G}^{\text{uni}}$ has a lower foot force than trotting and running is just due to the variance between runs as the maximum difference when not in a transition phase is $2$N.
Animals are observed to change gait to minimise actuator-structural forces (i.e. musculoskeletal forces) \cite{mechanicaltrigger}, which in biomechanics is measured through foot ground reaction force, $f_{\text{grf}}$. Similarly, $\pi_{G}^{\text{uni}}$ and $\pi_{G}^{\tau_{\%}}$ reduce $f_{\text{grf}}$ through selecting the optimal gait for minimising $\tau_{\%}$. This supports that $\tau_{\%}$ is a suitable alternative to $f_{\text{grf}}$, which is further validated by a strong correlation between them within Figure \ref{fig:metric_study}b. However, not only does $\pi_{G}^{\tau_{\%}}$ maintain a trotting gait past optimal $f_{\text{grf}}$, but also the transition itself is instantaneous. In turn, $\pi_{G}^{\tau_{\%}}$ better reflects the animal data from \cite{mechanicaltrigger} than $\pi_{G}^{\text{uni}}$. This could be explained by the metric being misinterpreted in \cite{mechanicaltrigger}; with a high correlation between $\tau_{\%}$ and $f_{\text{grf}}$ with stability metric $c_{\text{avg}}^{\text{err}}$, it suggests instability causes transition, which is rare on flat terrain, as addressed in Section \ref{sec:discussion}.

\subsubsection{Mechanical Work -- External Work}
Animals transition to preserve external mechanical work, $W_{\text{ext}}$, but do so before $\lambda^{W_{\text{ext}}}$, indicating relaxed minimisation (Figure \ref{fig:metric_study}a). $\pi_{G}^{\text{uni}}$ reflects this behaviour; transition occurs before $\lambda^{W_{\text{ext}}}$ yet minimal $W_{\text{ext}}$ is preserved. While $\pi_{G}^{W_{\text{ext}}}$ can reduce $W_{\text{ext}}$, its transition phase is extended over a larger range of velocities compared to animals, occurring just after $\lambda^{W_{\text{ext}}}$. In turn, this suggests that switching gaits between trotting and running offers minimal reductions in $W_{\text{ext}}$ resulting in a less definitive transition. However, $W_{\text{ext}}$ also seems to capture stability due to its high correlation with the stability metrics on $h_{\text{terr}}^{3}$ in Figure \ref{fig:metric_study}b, providing insight into its reduced role in Figure \ref{fig:metric_study}a where only flat terrain is present.

\subsubsection{Stability -- Stride Duration Coefficient of Variation}
Within \cite{stabTrans} animals are shown to reduce their stride duration coefficient of variation (CV) to preserve stability as high gait periodicity indicates stability. This behaviour is presented in Figure \ref{fig:metric_study}a, where animals are seen to initiate a transition phase when there is a considerable increase in stride CV, consequently leading to a decrease in CV and an increase in stability. $\pi_{G}^{\text{uni}}$ inherits the same strategy as only when a spike in stride CV is experienced does a transition phase begin resulting in improved stability through lowering stride CV. However, this is not the case with $\pi_{G}^{c^{\text{err}}}$ as it acts to reduce stride CV much more aggressively by mixing both slow, fast and auxiliary gaits which results in no clear transition phase being produced. 

Only $\pi_{G}^{\text{uni}}$ consistently reflects all animal data sets and demonstrates the successful instillation of animal gait transition strategies. This also supports the notion that no singular metric can characterise animal gait selection and only through unification can similar behaviour in robots arise; the minimisation of the metrics is expected and is seen across all $\pi_{G}$ policies, but the intricacies of animal gait transition strategies are only seen in $\pi_{G}^{\text{uni}}$. Additionally, we have verified this behaviour transfers to real-world deployment within Supplementary Section 2; $\pi_{G}^{\text{uni}}$ achieves an $18\%$ and $30\%$ reduction in CoT compared to trotting and running on grassy terrain while also preventing the failure cases trotting exhibits.

\begin{figure*}
    \centering
    \includegraphics[width=0.95\textwidth]{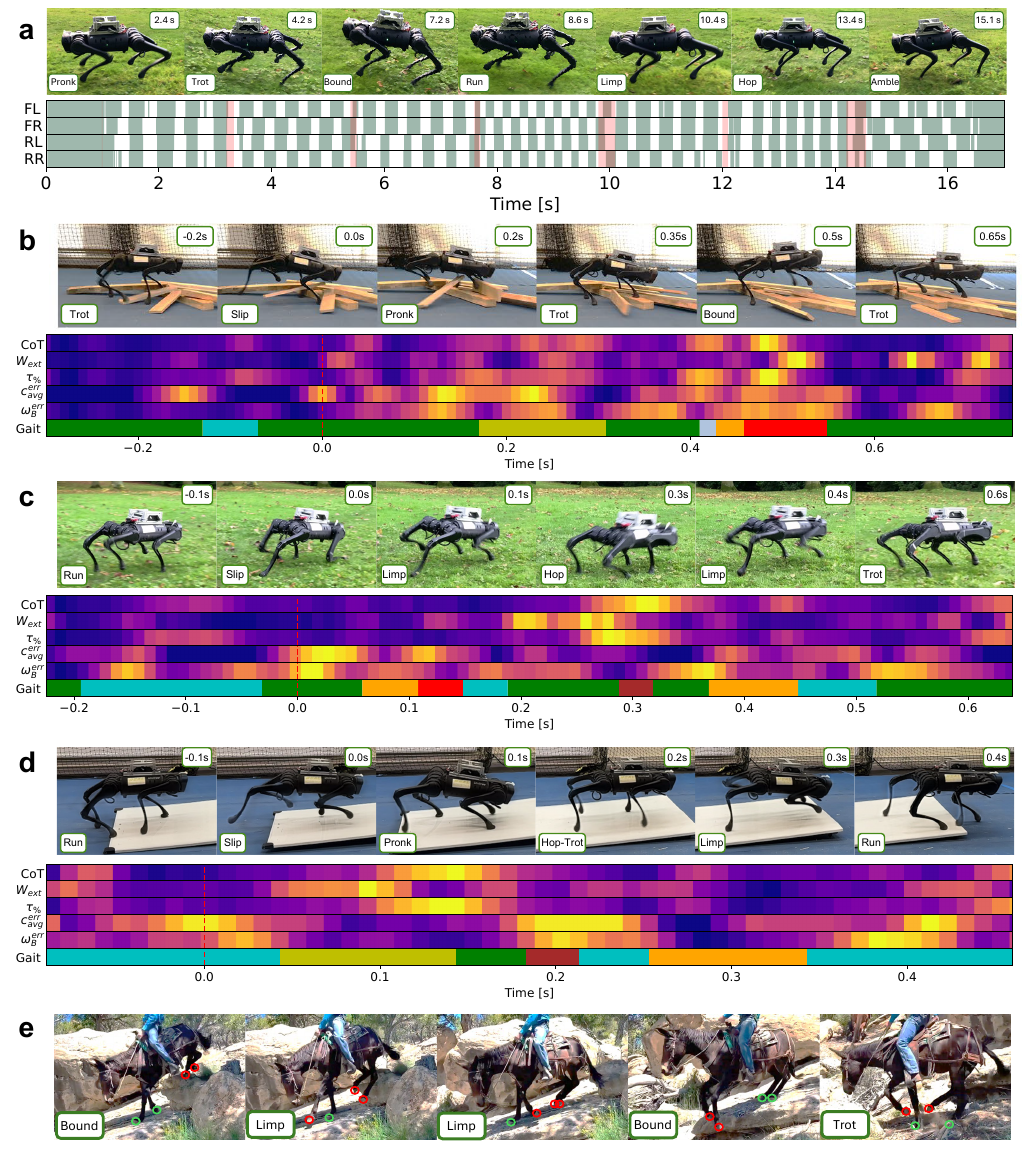}
    \centering
    \vspace{1mm}
    \caption{\textbf{Framework deployment in real-world environments to evaluate adaptability} (a) The deployment of $\pi_{L}^{\text{bio}}$ on uneven grassy terrain with manual gait selection to cycle through all gaits with a $v_{\text{x}}^{\text{cmd}}$ of $0.5$ m/s where red shaded regions indicate a transition period. For subfigures (b), (c) and (d) our framework is deployed on loose timber, muddy grass and a low-friction board, with maximum velocity magnitudes of $1$ m/s, $2$ m/s and $1.7$ m/s respectively, where the event of critical loss of stability is indicated by the red dashed line, the bottom subplot uses the same gait colour code as Figure \ref{fig:gs_policy_study}, and the heatmaps show the magnitudes of each metric increasing from purple to yellow. (e) Snapshots showing how animals also use a mixture of auxiliary gaits to overcome challenging terrain \cite{horsevid}, where red circles are swing feet and green circles are stance feet.}
    \label{fig:hw_experiments}
\end{figure*}

\subsection{Adaption to Real World Terrain}
\label{subsec:hw_exp}
%It has now been established that through applying a bio-inspired approach $\pi_{L}^{\text{bio}}$ exhibits proficient adaptability while $\pi_{G}^{\text{uni}}$ instills the framework with the same strategies animals use for gait selection, yet the question still remains: ultimately how proficient is this resultant framework when deployed within real world environments?
Although we have instilled animal gait transition strategies, gait procedural memory, and adaptive motion into our framework, its real-world proficiency without hardware deployment is uncertain. Grassy terrain may trap swing feet, and the ground often features irregularities. However, despite $\pi_{L}^{\text{bio}}$ only observing flat terrain during training, during hardware experiments it successfully realises all seven gaits on this terrain, as illustrated in Figure \ref{fig:hw_experiments}a, demonstrating that gait procedural memory and adaptive motion adjustment successfully transfer to real-world environments, providing a high level of adaptability, as shown in Supplementary Video S4.  

Terrain that causes states of instability presents a substantial risk to the robot, hence we test the limits of our framework through deployment on loose timber, muddy grass, and a low-friction board, as presented in Figure \ref{fig:hw_experiments}b, Figure \ref{fig:hw_experiments}c and Figure \ref{fig:hw_experiments}d respectively and consolidated in Supplementary Video S5, with an additional experiment for external perturbations included within Supplementary Section 3. Each of the presented experiments showcases an off-nominal stability recovery event; however in the nominal scenario, the framework can maintain stability without changing gaits. In the case of loose timber, critical instability is caused when one rear foot slips on a plank, causing it to collide with another. In response, $\pi_{G}^{\text{uni}}$ utilises auxiliary gaits pronk and bound to recover, as depicted in Figure \ref{fig:hw_experiments}b. This strategy of utilising the auxiliary gaits for stability recovery is seen across all experiments and reflected in animals as highlighted in Figure \ref{fig:hw_experiments}e where a horse is observed to utilise bounding and limping gaits to traverse down complex rock formations. 

In all experiments presented, a considerable increase in a combination of $W_{\text{ext}}$, $\tau_{\%}$ and $c_{\text{avg}}^{\text{err}}$ is experienced before a gait transition, while a weaker correlation is observed with CoT; spikes in $W_{\text{ext}}$, $\tau_{\%}$ and $c_{\text{avg}}^{\text{err}}$ directly coincide or even preempt gait changes while CoT peaks lag. This is expected with $c_{\text{avg}}^{\text{err}}$ and $W_{\text{ext}}$ as they capture periodicity and base height respectively. However, this is less expected for $\tau_{\%}$ to correlate with stability; this was never a factor investigated within \cite{mechanicaltrigger}, however, the correlation observed in these experiments and in Figure \ref{fig:metric_study}b provides strong evidence that this is the case. 

%%%%%%%%%%%%%%%%%%%%%%%%%%%%%%%%%%%%%%%%%%%%%%%%%%%%%%%%%%%%%%%%%%%%%%%%%%%%%%%%%%%%%%%%%
\section{Discussion}
\label{sec:discussion}
From taking inspiration from animal locomotion proficiency attributes, we have developed a locomotion framework capable of traversing complex and high-risk terrain despite the robot not utilising exteroceptive sensors nor experiencing any rough terrain during the training of $\pi_{L}^{\text{bio}}$. For $\pi_{L}^{\text{bio}}$, this is achieved through including the BGS output $\bm{\beta}_{L}$ (which encodes state dependent pseudo gait procedural memory and adaptive motion adjustments) within the observation space $\bm{o}_{L}$. This proves to be effective at overcoming rough terrain within Figure \ref{fig:loco_policy_study} as without the presence of $\bm{\beta}_{L}$ within $\bm{o}_{L}$, increased failure and instability are observed. This is the equivalent of removing an animal's cerebellum functionality (resulting in reduced limb coordination and stability \cite{cerebellarcontrol}) in turn supporting the claim that $\bm{\beta}_{L}$ effectively encodes pseudo gait procedural memory. 

$\pi_{G}^{\text{uni}}$ for optimal gait selection greatly expands adaptability through instilling gait selection strategies used by animals. As demonstrated in Figure \ref{fig:hw_experiments}, $\pi_{G}^{\text{uni}}$ can maintain stability in the event of the terrain undergoing radical structural or friction coefficient adjustments. These scenarios pose risks to robots with vision systems, as they typically cannot detect ground friction or terrain changes beyond their front legs. Through the use of $\pi_{G}^{\text{uni}}$, this limitation is mitigated without implementing resource-heavy exteroceptive sensors. Comparing Figure \ref{fig:hw_experiments}b through \ref{fig:hw_experiments}d to Figure \ref{fig:hw_experiments}e showcases that animals and $\pi_{G}^{\text{uni}}$ utilise multiple auxiliary gaits to prevent failure. This behavior, untargeted during training, suggests that unifying these metrics encodes the intricacies of animal gait transitions.

One provoking observation is that actuator-structural forces appear to characterise instability. Considering that \cite{mechanicaltrigger} validates by applying increased loads that could cause instability, it suggests this metric was initially misunderstood. Additionally, despite the employed biomechanics metrics only being tested on animals completing a linear forward trajectory on flat terrain, $\pi_{G}^{\text{uni}}$ upholds animal gait transition strategies across a wide range of terrains and base velocity commands. This supports that these metrics successfully characterise gait transitions and the notion that robots can indeed test biomechanics hypotheses, avoiding the resource, compatibility, and ethical challenges of animal testing. Moving forward, we aim to integrate our work with those that focus on Froude-free locomotion, such as \cite{parkour,deeptrackingcontrol,lifelike} to achieve exemplary adaptability and efficiency at both Froude-characterised and Froude-free locomotion levels.

%%%%%%%%%%%%%%%%%%%%%%%%%%%%%%%%%%%%%%%%%%%%%%%%%%%%%%%%%%%%%%%%%%%%%%%%%%%%%%%%%%%%%%%%%
\section{Methods}
\label{sec:methods}
\subsection{Control Framework Overview}
At the core of this work, the Unitree A1 quadruped robot used in all experiments features $12$ degrees of freedom, $n$, which are all modeled as revolute joints, with their angular positions denoted as $\bm{q}\in {\mathbb{R}}^{n}$ and its base orientation represented as a rotation matrix $\bm{R}_{B} \in SO(\text{3})$. As discussed in Section \ref{subsec:loco_bio_exp} and outlined in Figure \ref{fig:concept}, both $\pi_{L}$ and $\pi_{G}$ are integrated within a control framework and supported by the SE and BGS for generation of the robot's state data and gait references respectively.  The final output of $\pi_{L}$ is target joint positions, $\bm{q}^{*}$, which are converted into joint torques, $\bm{\tau}^{*}$, through the following proportional derivative controller that get sent to the motors,
\begin{align}
    \bm{\tau}^{*}=K_{p}(\bm{q}^{*}-\bm{q})-K_{d}\dot{\bm{q}},
\end{align}
where $K_{p}$ and $K_{p}$ are the proportional and derivative gains respectively. Throughout this work, a constant $K_{p}=25$ N/m and $K_{d}=1$ Ns/m are used while running at $1000$ Hz, while $\pi_{L}$ and $\pi_{G}$ are run at $500$ Hz and $100$ Hz respectively.

\subsection{Bio-inspired Gait Scheduler}
\label{subsec:BGS}
The BGS primary output, $\bm{\beta}_{L}=[\bm{c}^{\text{ref}}, \bm{p}_{x}^{\text{ref}},\bm{p}_{y}^{\text{ref}},\bm{p}_{z}^{\text{ref}}]\in {\mathbb{R}}^{\text{16}}$, defines the reference contact state of each foot, $\bm{c}^{\text{ref}} \in {\mathbb{B}}^{\text{4}}$, and their reference Cartesian position in the world frame $x$-axis, $\bm{p}_{x}^{\text{ref}}\in {\mathbb{R}}^{\text{4}}$, $y$-axis, $\bm{p}_{y}^{\text{ref}}\in {\mathbb{R}}^{\text{4}}$, and $z$-axis, $\bm{p}_{z}^{\text{ref}}\in {\mathbb{R}}^{\text{4}}$ which are calculated online using the Raibert heuristic \cite{raibert} to account for the current state of the robot. Throughout this paper, the limits enforced on the generation of $\bm{p}_{x}^{\text{ref}}$, $\bm{p}_{y}^{\text{ref}}$ and $\bm{p}_{z}^{\text{ref}}$ are $0.3$ m, $0.2$ m and $0.1$ m respectively from the nominal local foot position. An adjusted version of the BGS output, $\bm{\beta}_{G}$, is used for $\pi_{G}$ as not all the information in $\bm{\beta}_{L}$ is required. This has the form of $\bm{\beta}_{G}=[\bm{c}^{\text{ref}}, \bm{p}_{z}^{\text{ref}}, \Omega_{\text{stab}}, \kappa]\in {\mathbb{R}}^{\text{10}}$ where $\Omega_{\text{stab}}\in \mathbb{R}$ characterises the inherent stability of a gait \cite{Froude2}, and $\kappa \in \mathbb{B}$ is a logical flag to indicate a state of gait transition. We originally developed the BGS within \cite{joeybiogait} where the Froude number \cite{Froude2}, $\Omega$, is used to trigger gait transition based exclusively on CoT which results in a set order of transitions. However, when applied to this work this method is not entirely suitable as now multiple biomechanics metrics and a set of auxiliary gaits need to be considered. One issue is that $\Omega>1$ values are not compatible when calculating how many gait cycles, $C$, should a transition occur over. With this work investigating higher velocities than in \cite{joeybiogait}, this has been resolved through calculating $C$ through
\begin{align}
    C=e^{-2\Omega}
\end{align}
This relationship ensures an almost instantaneous transition at $\Omega\geq2$, which is the typical value that quadruped animals transition to a run \cite{Froude2} instantaneously. Another limitation is that the calculation of the transition resolution, $\delta$, (how quickly a transition should be progressed each time step) only enables the transition between set gait pairs; this was not an issue in \cite{joeybiogait} as CoT efficiency was the only metric considered. As $\pi_{G}$ requires any gait transition pair to be possible, $\Omega_{\text{stab}} =g/hf^{2}$ \cite{Froude2} is utilised, where $g$ is gravitational field strength, $h$ is hip height and $f$ is gait frequency. Through the use of $\Omega_{\text{stab}}$, we are able to determine an indication of the inherent stability of any gait, hence a transition between a higher $\Omega_{\text{stab}}$ gait to a lower one should have smaller values of $\delta$ to increase the smoothness of the transition to promote stability. In the reverse scenario, a more harsh transition is more feasible hence larger values of $\delta$ should be produced for rapid transition. As such, $\delta$ is now calculated by 
\begin{equation}
\label{eq:bgsresolution}
\delta=1+\frac{\Omega_{\text{stab}}}{\Omega_{\text{stab}}^{\text{next}}}
\end{equation}
where $\Omega_{\text{stab}}^{\text{next}}$ is the $\Omega_{\text{stab}}$ of the gait that's being transitioned to. In essence, $f$ of the current and next gait dictates the harshness of the transition. This behaviour is also reflected in animal gait transitions, where transitioning from running (higher $f$) to trotting (lower $f$) the transition is slower compared to the opposite scenario \cite{transspeed}. Overall, this augmented version of the BGS can achieve transition between any designed gait, while considering the inherent stability of the transition. Complete details of how $\bm{c}^{\text{ref}}$ is generated for each gait can be found in Supplementary Section 5.

\subsection{Policy Training}
\label{subsec:policytraining}
To simplify the training process, for both the locomotion policy, $\pi_{L}$, and gait selection policy, $\pi_{G}$, the training method, environment and network architecture are kept constant. Both policies are modelled as an MLP with hidden layer sizes $[512, 256, 128]$ and LeakyReLU activations. Subscripts $L$ and $G$ represent the specific parameters for the locomotion policy and gait selection policy respectively. The model-free DRL training problem for the policies is represented as a sequential Markov decision process (MDP), which aims to produce a policy that maximises the expected return of the policy $\pi$,
\begin{equation}
J(\pi) = \mathbb{E}_{\xi \sim p(\xi|\pi)} \left[ \sum_{t=0}^{N-1} \gamma^{t}r\right],
\end{equation}
in which $\gamma \in [0,1)$ is the discount factor, $\xi$ is a finite-horizon trajectory dependent on $\pi$ with length $N$, $p(\xi|\pi)$ is the likelihood of $\xi$, and $r$ is the reward function. The proximal policy optimization (PPO) algorithm \cite{ppo} is used to train all policies and the hyperparameters used are detailed in Supplementary Section 6, which were selecting through the standard method of parameter tuning. As discussed in Section \ref{subsec:loco_bio_exp}, we estimate the state of the robot during training through using an SE. Hence, in terms of applying state feedback noise for domain randomisation to improved sim-to-real transfer we only need to implement this on the input sensor data vector of the SE, $\bm{\sigma}=[\bm{\omega_{B}}, \dot{\bm{v}}_{B}, \bm{q}, \dot{\bm{q}}, \bm{\tau},\bm{f}_\text{grf}]$. This vector includes base angular velocity, $\bm{\omega_{B}}\in {\mathbb{R}}^{\text{3}}$, base linear acceleration, $\dot{\bm{v}}_{B}\in {\mathbb{R}}^{\text{3}}$, joint positions, $\bm{q}$, joint velocities, $\dot{\bm{q}}$, joint torques, $\bm{\tau}$, and foot ground reaction forces, $\bm{f}_\text{grf} \in {\mathbb{R}}^{\text{4}}$. As the initial state of the robot and its performance can never be guaranteed during real-world deployment, we also randomise the initial configuration of the robot, the mass of the robot's base, $K_{p}$ and $K_{d}$. Additionally, to ensure that a rich variation of $\bm{U}^{\text{cmd}}$ is experienced during training randomly sampled gaits, velocity commands and velocity change durations (to achieve random acceleration) are implemented. For all details regarding the noise and sampling used within training please refer to Supplementary Section 7. Although sim-to-real transfer can pose a considerable challenge when training DRL policies, we have found that through using domain randomisation, realistic and diverse velocity commands, and generating all robot state observations from the SE, our framework is able to achieve zero-shot traversal in all experiments and environments shown in Figures \ref{fig:terr_sc} and \ref{fig:hw_experiments}, hence demonstrating our methods sufficiently bridge the gap between simulation and the real world. The environment itself is constructed using RaiSim \cite{raisim}, as the vectorized environment setup allows for efficient training of policies. Additionally, the observation normalisation functionality offered by RaiSim is also used for improved training. 

During the training of $\pi_{L}$ only flat terrain is present within the environment to isolate and highlight the effect of implementing $\bm{\beta}_{L}$. A core claim of this work is that the implementation of $\bm{\beta}_{L}$ aims to impart gait procedural memory within $\pi_{L}^{\text{bio}}$ hence if rough terrain was observed during training it will become ambiguous if the improved performance is a direct result of implementing $\bm{\beta}_{L}$. However, for training $\pi_{G}$, flat to very rough terrain is implemented using fractal noise, enabling the policy to learn to employ the use of each gait minimising biomechanics metrics on a variety of terrains. We train all variations of $\pi_{L}$ and $\pi_{G}$ for $20$k iterations, taking 6 and 9 hours respectively, on a standard desktop computer with one Nvidia RTX3090 GPU with a training frequency of $100$Hz. It is also important to note that the training of all $\pi_{G}$ policies only utilise our final proposed bio-inspired locomotion framework $\pi_{L}^{\text{bio}}$.

\subsection{Locomotion Policy}
The goal of the locomotion policy $\pi_{L}$ is to realise the input $\bm{U}^{\text{cmd}}$ while exhibiting stable and versatile behaviour. As such, $\pi_{L}$ is trained to generate the action, $\bm{q}^{*}$, from an input observation, $\bm{o}_{L} = [\bm{\beta}_{L}, \bm{s}, \bm{v}_{B}^{\text{cmd}}]\in {\mathbb{R}}^{\text{69}}$, where $\bm{v}_{B}^{\text{cmd}}=[v_{x}^{\text{cmd}}, v_{y}^{\text{cmd}}, \omega_{z}^{\text{cmd}}] \in {\mathbb{R}}^{\text{3}}$ is the high-level velocity command of the robot's base within $\bm{U}^{\text{cmd}}$, as outlined in Figure \ref{fig:concept}. $\bm{s}$ is generated from the output of the SE and is defined as $\bm{s} = [\bm{\alpha} \bm{R}_{B}^{T}, \bm{q}, \bm{\omega}_{B}, \dot{\bm{q}}, \bm{v}_{B}, z_{B}, \bm{\tau}, \bm{c}] \in {\mathbb{R}}^{\text{50}}$, where $\bm{\alpha}=[0,0,1]^{T}$ is used to select the vertical $z$-axis, $\bm{\omega}_{B}\in {\mathbb{R}}^{\text{3}}$ is the base angular velocity, $\bm{v}_{B}\in {\mathbb{R}}^{\text{3}}$ is the base linear velocity, $z_{B}$ is the current base height, and $\bm{c} \in {\mathbb{B}}^{\text{4}}$ is the contact state of the feet. The locomotion reward function, $r_{L}$, is formulated so that the the output of the policy can realise the reference gait patterns and velocity commands stably, smoothly and accurately, 
\begin{equation}
r_{L}=\text{w}_{\eta}r_{\eta}+\text{w}_{v^{\text{cmd}}}r_{v^{\text{cmd}}}+\text{w}_{f}r_{f}+\text{w}_{\text{stab}}r_{\text{stab}},
\end{equation}
where $r_{\eta}$, $r_{v^{\text{cmd}}}$, $r_{f}$ and $r_{\text{stab}}$ are the grouped reward terms focusing on efficiency, velocity command tracking, gait reference tracking and stability respectively. $\text{w}_{\eta}$, $\text{w}_{v^{\text{cmd}}}$, $\text{w}_{f}$ and $\text{w}_{\text{stab}}$ are the weights of each reward and are valued at $-1.5$, $15$, $-10$, and $-5$ respectively. $r_{\eta}$ aims to minimise joint jerk, $\dddot{\bm{q}}$, joint torque, and the difference between $\bm{q}^{*}$ and the previous action, $\bm{q}^{*}_{t-1}$, 
\begin{equation}
    r_{\eta}=\|\dddot{\bm{q}}\|^{2}+\|\bm{\tau}\|^{2} + \|\bm{q}^{*}-\bm{q}^{*}_{t-1}\|
\end{equation}
$r_{v^{\text{cmd}}}$ minimises the difference between the commanded base velocity and the current base velocity,
\begin{equation}
\label{eq:veltrackrew}
    r_{v^{\text{cmd}}}=\psi \left( \|\bm{v}_{B}-\bm{v}_{B}^{\text{cmd}}\|^{2} \right),
\end{equation}
in which the function $\psi:x \to 1-\tanh \left(x^{2} \right)$ is used to normalise the reward term so that their maximum value is $1$ to prevent bias towards individual rewards, $\bm{v}_{B}=[v_{x}, v_{y}, \omega_{z}] \in {\mathbb{R}}^{\text{3}}$ is the current base $x$, $y$ and yaw velocities. $r_{f}$ ensures the robot realises the commanded gait references within $\bm{\beta}_{L}$,
\begin{equation}
    r_{f}=|\bm{c}^{\text{err}}|+\sum_{i=1}^{4} \|\bm{p}_{i}-\bm{p}_{i}^{\text{ref}}\|^{2},
\end{equation}
where $\bm{c}^{\text{err}} \in {\mathbb{B}}^{\text{4}}$ defines the feet that do not meet the desired contact state, with $\bm{p}_{i}\in {\mathbb{R}}^{\text{3}}$ and $\bm{p}_{i}^{\text{ref}}\in {\mathbb{R}}^{\text{3}}$ being the current and reference Cartesian positions of the $i$-th foot. $r_{\text{stab}}$ aims to prevent contact foot slip, large hip joint motions and undesirable base orientations,
\begin{equation}
\begin{split}
    r_{\text{stab}}=\sum_{i=1}^{F} \|\dot{\bm{p}_{i}}\|^{2} + \|\bm{\omega}_{B}\|^{2} + \psi\left(\|\bm{\alpha} \bm{R}_{B} - \bm{\alpha}\bm{R}_{B}^{\text{des}} \|^{2}\right) \\
    - \psi \left(\left(z_{B}-z_{B}^{\text{nom}}\right)^{2} \right)+ \| \bm{q}_{\text{hip}} \|^{2},
\end{split}
\end{equation}
where $\bm{p}_{i}\in {\mathbb{R}}^{\text{3}}$ is the velocity of the $i$-th foot scheduled to be in stance, $F$ is the number of stance feet, $\bm{\omega}_{B}=[\omega_{x}, \omega_{y}] \in {\mathbb{R}}^{\text{2}}$ where $\omega_{x}$ and $\omega_{y}$ are roll and pitch base velocities respectively, $\bm{R}_{B}^{\text{des}} \in SO(\text{3})$ is the desired base orientation, $z_{B}^{\text{nom}}$ is the nominal base heights respectively, and $\bm{q}_{\text{hip}} \in {\mathbb{R}}^{\text{4}}$ is the hip angular joint positions. Overall, this reward function enables deployment of all targeted gaits with rapid transitions between them, even at high speeds, as demonstrated in Figures \ref{fig:loco_policy_study}, \ref{fig:gs_policy_study} and \ref{fig:hw_experiments}.

\subsection{Biomechanics Gait Transition Metrics}
\label{subsec:biometrics}
%\hl{we estimate CoT and torque limits and justify}
Although the set of biomechanics metrics applied in this work were designed to accommodate different animals of the same morphology, even when animal body size and weight vary considerably, the fact still remains that they were designed for the analysis of animal locomotion. Hence, several adjustments to how they are calculated needs to be implemented; for example, energy consumption in animals is often measured through the rate of consumption of $\text{O}_{2}$, hence unsuitable for the application of robotics. Additionally, as robots provide a wide array of feedback data some of the metrics have also been augmented to better reflect the characteristic that these biomechanics metrics are attempting to characterise. That being said, for Figure \ref{fig:metric_study}a only the original biomechanics metrics are applied to allow for direct comparison between robot and animal data.

\subsubsection{Energy Efficiency}
%\hl{add point of estimating the CoT}
The calculation of CoT takes the general form of 
\begin{align}
    \text{CoT} = \frac{P}{mgv},
\end{align}
where $P$ is power consumed and $m$ is the system's mass. When studying animal locomotion, $P$ is found through measuring how much $\text{CO}_{2}$ is generated and $\text{O}_{2}$ is consumed and $v$ is assumed to be the speed of the treadmill the animal is running on \cite{CoTTrans, CoTTrans2, Froude2}. For the case of the robot, we calculate $P$ from $\bm{\tau}$ and $\dot{\bm{q}}$ with an adjustment term, adopted from \cite{RLgaitgen}, and $v$ is assumed to be the magnitude of the robot base velocity command to take a similar approach to animal studies and for consistent metric use between simulation and real-world deplopyment; completely accurate measurement of the robot's linear base velocity is impossible during real-world deployment due to the accumulation of error within the SE. As such, calculation of the robot's CoT is formulated as
\begin{align}
    \text{CoT} = \sum_{i=1}^{n} \frac{\max(\tau_{i}\dot{q}_{i}+0.3\tau_{i}^{2},0)}{mg|\bm{v}_{B}|},
\end{align}
where $m$ is the robot's mass and $g$ is gravity. It should be noted that CoT is only calculated and applicable when $|\bm{v}_{B}^{\text{cmd}}|>0$.

\subsubsection{Actuator-structural Forces}
As gaining an exact understanding of the actuator-structural forces within animals is infeasible, researchers have opted instead to measure the peak ground reaction forces of the animal's stance feet during locomotion using force plates \cite{mechanicaltrigger}. Other methods include adding strain gauges to the bones of the animals \cite{mechanicaltrigger3}. However, in the case of robots we have the privilege of having access to joint state feedback while also knowing the exact limitations of the hardware. Therefore, when considering the biomechanics hypothesis that animals aim to minimise actuator-structural forces to prevent injury and that torque is proportional to strain and force, in the case of the robot we chose to characterise the actuator-structural forces through joint torque saturation, $\tau_{\%}$, which is calculated by
\begin{align}
    \tau_{\%}=\left| \frac{\bm{\tau}}{\bm{\tau}_{\text{lim}}} \right| \frac{1}{n},
\end{align}
where $\bm{\tau}_{\text{lim}} \in {\mathbb{R}}^{n}$ is the joint torque limits (assumed based on manufacturers specification), which proves particularly usefully when considering that the hip joints of most quadruped robots, including the A1, are often more sensitive to forces at the foot due to their distance from the point of ground impact and the only motor of the leg set in this plane; this would not be considered if just ground reaction force was used to characterise actuator-structural forces. 

\subsubsection{Mechanical Work Efficiency}
During animal locomotion, if they are to have perfect mechanical work efficiency there would be a net zero change in external work over the duration of a gait cycle as there would be perfect exchange between kinetic and potential energy \cite{mechwork}. As expected perfect mechanical work is never seen in nature, hence mechanical work efficiency in animals is characterised by the sum of the change in kinetic and potential energy \cite{mechwork} or the sum of the external work of the animal \cite{mechwork2} over the duration of a gait cycle. As this is typically calculated through measuring the $\text{O}_{2}$ uptake, for the case of robots we formulate the calculation of the external work, $W_{\text{ext}}$, through
\begin{align}
    W_{\text{ext}}=\sum_{i=0}^{t_{\text{gait}}} \left( \Delta E_{\text{k},i}-\Delta E_{\text{p},i} \right)
\end{align}
where $t_{\text{gait}}$ is the duration of the current gait cycle, and $\Delta E_{\text{k},i}$ and $\Delta E_{\text{p},i}$ are the changes in kinetic and potential energy over a control time step respectively. The primary difference between the metrics seen in biomechancis and our formulation of $W_{\text{ext}}$ is that $\Delta E_{\text{k},i}$ accounts for not only forward linear velocity but also lateral and angular velocity whereas originally only forward linear velocity is considered. 

\subsubsection{Stability}
The best indication of stability in animals is their stride duration CV. This metric characterises periodicity, which is a primary indication of stable locomotion \cite{stabTrans}. However, to accurately calculate this, the mean and standard deviation of the stride duration needs to be taken over an extended period of time for appropriate data generation. This is sufficient for undertaking analysis similar to that presented in Figure \ref{fig:metric_study}a, but this presents an issue when it comes to analysing the performance of the proposed control framework as it is common for multiple speed commands being used within the duration of one stride. Hence, to overcome this limitation we instead use $c_{\text{avg}}^{\text{err}}=|\bm{c}^{\text{err}}|/4$, which can be measured every time step rather than just at each foot touchdown event; the gait references generated by the BGS have a constant and periodic stride duration, therefore an accurate tracking of this reference would in turn indicate high periodicity, which is further supported by the correlation between the two metrics within Figure \ref{fig:metric_study}b. 

\subsection{Gait Selection Policy}
\label{subsec:gaitselectionpolicy}
To achieve optimal gait selection for a given state, we leverage the biomechanics metrics within the reward function of $\pi_{G}$, $r_{G}$. For the different variations of $\pi_{G}$ used within Figure \ref{fig:metric_study}a, each policy's reward function only features the metric that its focusing on within $r_{G}$ but $\pi_{G}^{\text{uni}}$ unifies all metrics hence uses the full form of $r_{G}$ with all metrics. In addition, as the biomechanics metrics all describe characteristics that animals try to minimise through changing gaits, they can be directly applied within $r_{G}$ with some normalisation where appropriate. The full form of $r_{G}$ is
\begin{equation}
    r_{G}=\text{w}_{\text{u}}r_{\text{u}}+\psi(\text{CoT})+\psi(\tau_{\%})+\psi(c_{\text{avg}}^{\text{err}})+\psi(W_{\text{ext}}),
\end{equation}
where $r_{\text{u}}$ is the utility reward term which all $\pi_{G}$ use and $\text{w}_{\text{u}}$ is its weight with a value of $0.4$. $r_{\text{u}}$ aims to ensure the smoothness of the output $\Gamma^{*}$, the standing gait is only used when appropriate, and any select gait is still able to follow $\bm{v}_{B}^{\text{cmd}}$. To achieve this, $r_{\text{u}}$ has the form of 
\begin{align}
    r_{\text{u}}=r_{v^{\text{cmd}}}+r_{\text{stand}}+r_{\text{smooth}},
\end{align}
in which $r_{v^{\text{cmd}}}$ is taken from \eqref{eq:veltrackrew}, and $r_{\text{stand}}$ is set to $-10$ if a stand gait is used when $|\bm{v}_{B}^{\text{cmd}}|>0$ or not used when $|\bm{v}_{B}^{\text{cmd}}|=0$. For $r_{\text{smooth}}$, the reward aims to penalise unnecessary changes in $\Gamma^{*}$ to remove rapid gait changes when two gaits could achieve similar metric minimisation for a given task and state. As such, if there is a gait change between time steps it is calculated as $r_{\text{smooth}}=-\psi(\text{CoT}+\tau_{\%}+c_{\text{avg}}^{\text{err}}+W_{\text{ext}})$ otherwise it is set to $0$. To generate $\Gamma^{*}$, $\pi_{G}$ takes in input observation vector $\bm{o}_{G}=[\bm{s}, \bm{\beta}_{G}, \bm{v}_{B}^{\text{cmd}}, \dot{\bm{v}}_{B}^{\text{cmd}}, \Gamma^{*}_{t-1}]\in {\mathbb{R}}^{\text{66}}$ in which $\Gamma^{*}_{t-1}$ is the previous output action to aid in action smoothing. Appropriate selection of the data provided to $\pi_{G}$ is critical in order to achieve targeted minimisation of the biomechanics metrics. As such, the inclusion of $\bm{s}$ coupled with $\bm{c}^{\text{ref}}$, $\bm{p}_{z}^{\text{ref}}$, $\bm{v}_{B}^{\text{cmd}}$, $\dot{\bm{v}}_{B}^{\text{cmd}}$ and $\Omega_{\text{stab}}$ informs the policy of its current and demanded stability, while the terms $\bm{\tau}$ and $\dot{\bm{q}}$ within $\bm{s}$ capture the power consumption of the robot and the forces to which it is subjected. Overall, through the formulation of the biomechanics metrics within this reward function, we are not only able to full investigate the effects in gait selection when minimising each metric, but also instil the intrinsics of animal gait transition strategies within a DRL gait selection policy, as detailed in Figure \ref{fig:metric_study}.

%%%%%%%%%%%%%%%%%%%%%%%%%%%%%%%%%%%%%%%%%%%%%%%%%%%%%%%%%%%%%%%%%%%%%%%%%%%%%%%%%%%%%%%%%

\section*{Data Availability}
The simulation and hardware experimental data used to generate the figures within this paper are available at \url{https://github.com/ihcr/learning_to_adapt} \cite{ltagitrepo}. The animal data within the repository was extracted from \cite{stabTrans, Froude1, mechwork2}, \cite{mechanicaltrigger} and \cite{mechwork2}.

\section*{Code Availability}
The code and demos of our framework is available on GitHub at \url{https://github.com/ihcr/learning_to_adapt} \cite{ltagitrepo}.

\section*{Code Availability}
The code and demos of our framework is available on GitHub at \url{https://github.com/ihcr/learning_to_adapt} \cite{ltagitrepo}.

\section*{Acknowledgments}
This work was partially supported by the Royal Society [grant number RG\textbackslash R2\textbackslash232409] and the Advanced Research and Innovation Agency [grant number SMRB-SE01-P06].

\section*{Author Contributions Statement}
J.H.: conceptualisation, formal analysis, control framework design, software design, simulation design, hardware experiment design, data collection and analysis, wrote the manuscript;
C.Z.: conceptulisation, formal analysis, discussion, review and editing, supervision, funding acquisition.

%%%%%%%%%%%%%%%%%%%%%%%%%%%%%%%%%%%%%%%%%%%%%%%%%%%%%%%%%%%%%%%%%%%%%%%%%%%%%%%%%%%%%%%%%

\bibliographystyle{IEEEtran}
\bibliography{Bio_bib_NMI}
\newpage
\beginsupplement
\onecolumn
This supplementary information includes:

Sections S1-S7

Tables S1-S4

Figures S1 and S2

Movies S1-S5

\section{Classification of Froude-characterised and Froude-free Locomotion}
Terrestrial locomotion can take on many forms but when it comes to animal quadruped locomotion two main groups can be identified, locomotion that can be characterised by the Froude number \cite{Froude2} and locomotion that cannot:
\begin{itemize}
    \item Froude-characterised locomotion
    \begin{itemize}
        \item This includes walking, running, hopping, ambling, pronking, pacing, bounding, limping, cantering, galloping and other rhythmic gaits.
        \item Can account for $\approx70-90\%$ of the daily locomotion utilised by terrestrial animals \cite{locopercent1,locopercent2}.
        \item Gait transition need to occur at both low and high velocity magnitudes, hence requiring the ability to rapidly change gait under dynamic conditions.  
        \item Features a desired velocity that has not normal component to the ground plane.
    \end{itemize}
    \item Froude-free locomotion
    \begin{itemize}
        \item This includes climbing, jumping, crouching and other non-rhythmic gaits or motion skills.
        \item Can account for $\approx10-30\%$ of the daily locomotion utilised by terrestrial animals \cite{locopercent1,locopercent2}.
        \item Gait and skill transitions occur at low velocity magnitudes due to the extreme differences in motion.
        \item Features a desired velocity that has normal components to the ground plane.
        \item Breaks the assumptions that are used in the formulation of the Froude number due to large variations in the height of the base relative to the ground.
    \end{itemize}
\end{itemize}
With Froude-characterised locomotion being the primary locomotion utilised by animals to traverse natural terrain, this exemplifies the importance of quadruped robots mastering this skill for optimal performance in the majority of scenarios. Consequently, it is this highly adaptive gait-based locomotion and transition strategies seen in animals that we aimed to instil within our locomotion framework for optimal Froude-characterised locomotion. 
\newpage
\section{Preservation of Biomechanics Metrics on Grass}
\label{sup:gssprint}
When deploying $\pi_{L}^{\text{bio}}$ with $\pi_{G}^{\text{uni}}$ on a smooth low-friction floor and uneven grassy terrain $\pi_{G}^{\text{uni}}$ can utilise the same behaviour animals demonstrate in minimising CoT; comparing the performance of maintaining a static gait of both trot and run gaits within Supplementary Figure 1, $\pi_{G}^{\text{uni}}$ can select the most energy efficient gait and even prevent failure as the trot gait cannot maintain stability for the entire experiment. Additionally, the distribution of gait usage between trot and run gaits varies between the two terrains. This is due to the uneven grassy terrain causing reduced stability of the robot when trotting, as indicated by $c_{\text{avg}}^{\text{err}}$ in Supplementary Figure 1, which in response $\pi_{G}^{\text{uni}}$ exhibits increased modulation between trotting and running to preserve stability. This preservation of stability while minimising CoT is achieved through the resultant modulation of the stride frequency as gaits with higher stride frequency offer improved stability \cite{stabTrans} and efficiency on rough terrain \cite{mechwork2}. Overall, on smooth low-friction floor $\pi_{G}^{\text{uni}}$ is able to achieve a $15\%$ and $24\%$ reduction in CoT compared to trotting and running gaits and a reduction of $18\%$ and $30\%$ on uneven grass.

\begin{figure}[h]
    \centering
    \includegraphics[width=1.0\textwidth]{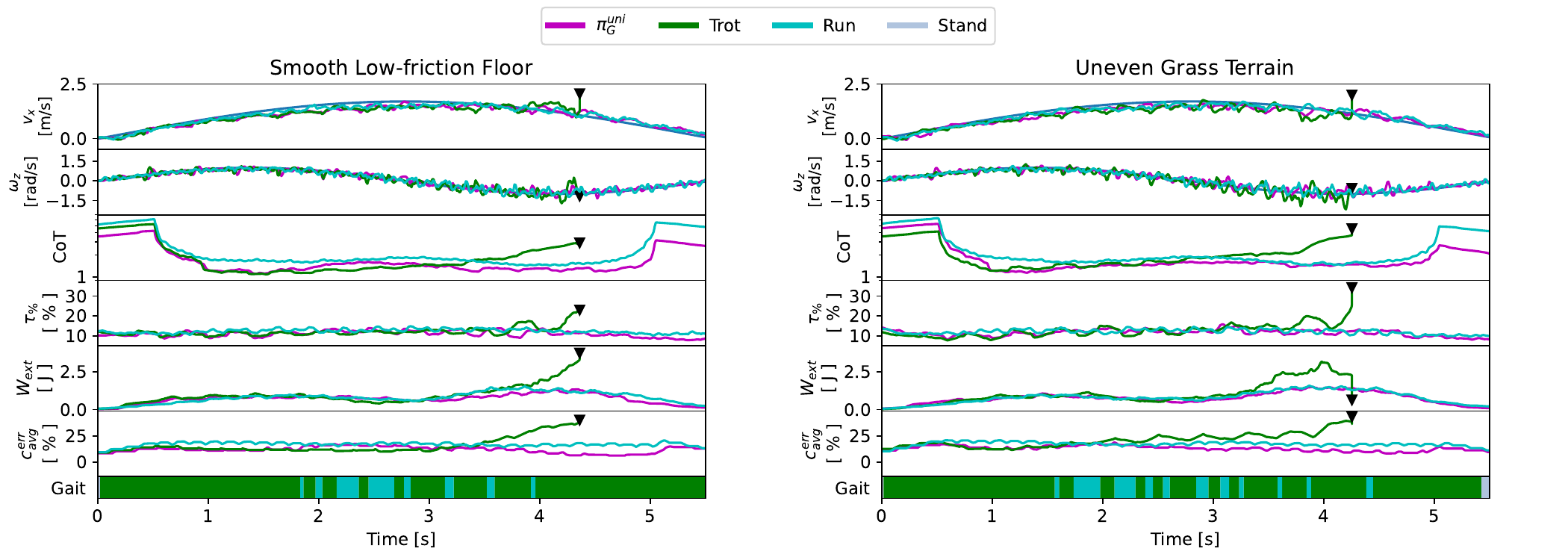}
    \vspace{-2mm}
    \caption{Deployment of $\pi_{L}^{\text{bio}}$ and $\pi_{G}^{\text{uni}}$ on smooth low-friction floor and uneven grassy terrain, with the black triangles indicating a point of failure.}
    \label{fig:gs_sprint}
\end{figure}

\section{Stability Recovery from External Perturbations}
\label{sup:recoverybroom}

An additional stability recovery experiment was completed where the robot was subjected to repeated external perturbations and in response utilised the auxiliary gaits to recover, as presented in Supplementary Figure 2, which in turn reflects the findings discussed in Sections \ref{subsec:hw_exp} and \ref{sec:discussion}.

\begin{figure}[h]
    \centering
    \hspace{5mm}
    \includegraphics[width=0.8\textwidth]{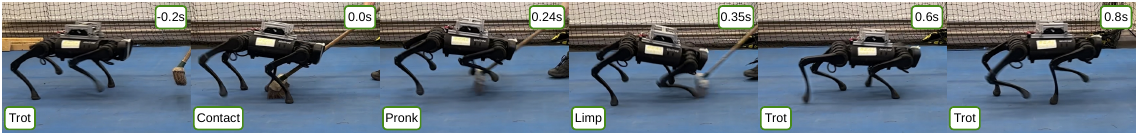}
    \includegraphics[width=1\textwidth]{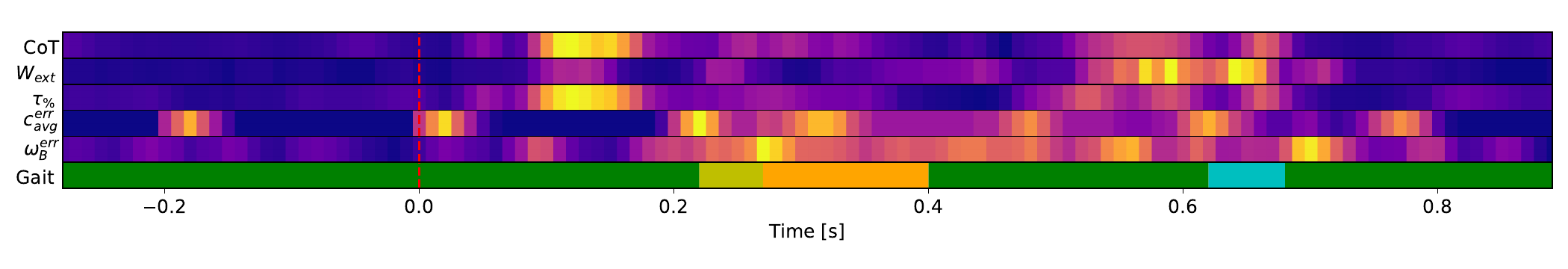}
    \vspace{-5mm}
    \caption{Enacting perturbations upon the legs of the robot to investigate how our framework can reject this instability.}
    \label{fig:gs_recovery_broom}
\end{figure}
\newpage
\section{Comparison to Other Froude-characterised Locomotion Frameworks}
\subsection{Performance and Characteristics Comparison}
To benchmark the performance of our framework, we have compiled all available performance metrics and characteristics of other similar frameworks in Supplementary Table 1, with the most relevant ones being:
\begin{itemize}
    \item DeepTransition \cite{natureCommunications}
    \item WalkTheseWays \cite{RLmanyrefs}
    \item RL+MPC \cite{RLmpcref}
    \item Phase-guided \cite{RLe2eGaitTrans}
    \item EnergyLocomotion \cite{RLenergymin}
\end{itemize}
Comparing our framework to the others that take a bio-inspired approach, DeepTransition and EnergyLocomotion, it has been deployed in a much wider variety of terrains. In respect of DeepTransition, this disparity could be due to their limited number of deployable gaits and their method of utilising a bio-inspired approach at the low-level, which as discussed in the main text, would limit its performance even though both our methods and theirs take inspartation from stability and efficiency animal gait strategies. In the case of EnergyLocomotion, as only one energy efficiency animal gait strategy was implemented, only three gaits emerged which in turn results in limited adaptability in comparison to our framework.

\begin{table}[h]
\centering
\caption{Comparison between Froude-characterised Locomotion Frameworks} \label{tab:fw_comp}
\vspace{-1mm}
\begin{tabular}{lcccccc}
\toprule
\multicolumn{1}{l}{\parbox{1cm}{\centering }} & \multicolumn{1}{l}{\parbox{2cm}{\centering Phase-guided}} & \multicolumn{1}{l}{\parbox{2.cm}{\centering WalkTheseWays}} &  \multicolumn{1}{l}{\parbox{2cm}{\centering RL+MPC}} & \multicolumn{1}{l}{\parbox{1cm}{\centering DeepTransition}} & \multicolumn{1}{l}{\parbox{2cm}{\centering EnergyLocomotion}} & \multicolumn{1}{l}{\parbox{3cm}{\centering \textbf{Ours}}} \\ \vspace{-2mm} \\
\midrule 
Framework & CPG-RL & RL & RL-MPC & CPG-RL-IK & RL & RL \\
Type &  &  &  &  &  & \\
 &  Amble, & Pronk, & Trot, Pace & Amble, & Pronk & Amble, Run\\
Gaits & Pace, Trot, & Pace, Trot, & Pronk, Bound & Trot, Pronk & Amble, Trot,  & Pronk, Bound\\
 & Bound & Bound & Gallop & Bound &  & Limp, Amble, Hop \\
 &  &  &  &  &  & \\
 Level of &  &  &  & Low-Level & High-level for & High-level for \\
 Bio-inspired  & \xmark & \xmark & \xmark & Stability and & Efficiency & Stability and \\
 Implementation &  &  &  & Efficiency &  & Efficiency \\
 &  &  &  &  &  & \\
Optimal Gait  & \xmark & \xmark & \xmark & \cmark & \cmark & \cmark \\
Selection &  &  &  &  &  & \\
 &  &  &  &  &  & \\
Arbitrary Gait  & \cmark & \cmark &\cmark & \xmark & \xmark & \cmark \\
Transitions &  &  &  &  &  & \\
 &  &  &  &  &  & \\
Hardware & Black Panther & Go1 & Go1 & A1 & A1 & A1 \\
 &  &  &  &  &  & \\
 &  &  &  &  &  & Wood-chip, Deep Rocks, \\
 &  &  &  &  & & Large Step, Concrete Slabs, \\
Terrains & Grass  & Sand, Steps, & Grass,Tarmac, & Flat Terrain & Cardboard Sheets & Tarmac, Loose Timber, \\
Traversed &  & Grass, Wood-chip & Wooden Boards & with Large Gaps & Grass, Rocks, Bush, & Overgrown Roots, Leaves,\\
 &  &  &  &  & Wooden Planks & Grass, Balanced Timber, \\
 &  &  &  &  &  & Low-friction Ramp \\
\bottomrule
\end{tabular}
\vspace{-2mm}
\end{table}

Additionally, both these frameworks cannot achieve arbitrary gait transitions as their gait selection strategies are embedded within their locomotion method, while our framework is able to achieve this due to $\pi_{L}^{\text{bio}}$ and $\pi_{G}^{\text{uni}}$ being independent policies which in turn promotes its suitability to be combined with Froude-free locomotion frameworks, as will be discussed in the next section. In the comparison between our framework and those who haven't implemented neither optimal gait selection nor implemented any animal gait strategies (WalkTheseWays, RL+MPC and Phase-guided), with our framework being able to successfully traversing a much more diverse set of terrains and realise more types of gaits, this supports the notion that our method provides considerable improvements in adaptability and proficiency in Froude-characterised locomotion.
\newline
\subsection{Comparison Between Froude-characterised to Froude-free Focused Frameworks}
Comparison between Froude-characterised focused to Froude-free focused locomotion frameworks is somewhat impractical. Primarily, Froude-free locomotion frameworks measure success through failure rate over challenging terrain and stability, with no concern or analysis on efficiency. Additionally, these frameworks are only able to realise a walking or running gait, with other Froude-characterised gaits and optimal gait selection being outside their operational scope. On the other hand, our framework, that focuses on Froude-characterised locomotion, determines its success based on command tracking error, rapid but stable gait transitions, efficiency and adaptability. Consequently, direct comparison between these two locomotion groups is not constructive as they are inherently different in terms of the challenges they are designed to overcome. However, an important and valuable direction for further research would be investigating how to effectively combine both framework types.

For example, although the Froude-free locomotion could potentially overcome the terrains featured in Figure 6 of the main text, they would do so without minimising energy consumption, with increased energy expenditure likely to be caused by excessive motion; as we have proven that the terrains in Figure 6 are traversable through just switching Froude-characterised gaits, use of Froude-free skills like climbing, jumping and sure-footedness would incur unreasonable energy costs. 
However, as we explicitly focus on optimal Froude-characterised locomotion in our work, through instilling animal gait strategies, our framework is not only able to traverse these terrains while preserving efficiency through optimal gait selection, but also recover from critically unstable states without requiring the use of inefficient Froude-free locomotion through rapid gait transitions to auxiliary gaits and then back to a nominal gait in a matter of milliseconds, as exhibited in Figure 6.

\section{Generation Gait Contact References within the BGS}
\label{sup:gaitgen}
\begin{table}[h]
\centering
\caption{Gait design parameters.} \label{tab:gait_params}
\vspace{-1mm}
\begin{tabular}{lcccl}
\toprule
\multicolumn{1}{l}{\parbox{1cm}{\centering Name}} & \multicolumn{1}{l}{\parbox{1cm}{Period}} & \multicolumn{1}{l}{\parbox{1.cm}{\centering Duty Factor}} &  \multicolumn{1}{l}{\parbox{2cm}{\centering Phase Offset}}  \\ \vspace{-2mm} \\
\midrule 
Trot & 0.40 & 0.50 & 0.00, 0.50, 0.50, 0.00 \\
Run & 0.30 & 0.40 & 0.00, 0.50, 0.50, 0.00 \\
Bound & 0.40 & 0.40 & 0.00, 0.00, 0.50, 0.50 \\
Pronk & 0.50 & 0.50 & 0.00, 0.00, 0.00, 0.00 \\
Amble & 0.50 & 0.55 & 0.00, 0.50, 0.25, 0.75 \\
Unnatural & 0.40 & 0.50 & 0.05, 0.50, 0.50, 0.00 \\
Hop & 0.30 & 0.50 & 0.00, 0.00, 0.00, 0.00 \\
\bottomrule
\end{tabular}
\vspace{-2mm}
\end{table}
In accordance to our work in \cite{joeybiogait}, to generate $\bm{c}^{\text{ref}}$, a set of phase variables, $\phi_{i} \in [0, 1)$, for each leg $\phi_{1}, \ldots, \phi_{4}$, are used to determine the progress along a gait pattern, and duty factor, $d_{i} \in [0,1)$, sets the percentage of the phase that each leg is in stance. These encoded parameters for each gait are presented within Supplementary Table 2. When $\phi_{i} = d_{i}$, the contact state of the $i$-th leg switches to swing. The phase of the $i$-th leg is calculated as 
\begin{equation}
\label{phase_update}
\phi_{i} = \frac{t-t_{i,0}}{T},
\end{equation}
in which $t$ is the current time, $t_{i,0}$ is the start time of the current gait period of the $i$-th leg, and $T$ is the gait period. The last parameter used to construct a gait pattern is the phase offset, $\theta_{i} \in [0, 1]$, that defines the difference in phase between the leading leg and all other legs through $\phi_{i} = \phi_{1} + \theta_{i}$. As such, dependent on the phase and gait parameters, the $i$-th value of $\bm{c}^{\text{ref}}$ is either $1$ or $0$ to signify if the leg should be in stance or swing respectively.

\section{PPO Hyperparameters}
\label{sup:hyperparameters}
All PPO hyperparameters used within this work are detailed within Supplementary Table 3.
\begin{table}[h]
\centering
\caption{PPO Hyperparameters.} \label{tab:ppo_params}
\vspace{-1mm}
\begin{tabular}{lclc}
\toprule
\multicolumn{1}{l}{\parbox{1cm}{\centering Parameter}} & \multicolumn{1}{l}{\parbox{1cm}{\centering Value}}   \\ \vspace{-2mm} \\
\midrule 
Number of Environments & 240 \\ 
Clip Range & 0.2 \\
Max Steps per Batch & 400 \\
GAE $\lambda$ & 0.95 \\
Learning Epochs per Batch & 4 \\ 
Learning Rate & 5e-4 \\
Number of Mini-batches & 4 \\
Minimum Policy std & 0.2 \\
Reward Discount Factor & 0.99 \\
Optimizer & Adam \\
\bottomrule
\end{tabular}
\vspace{-2mm}
\end{table}
\newpage
\section{Noise and Sampling Distributions Used During Training}
\label{sup:distributions}
In the effort of improving sim-to-real transfer, domain randomisation was utilised through randomly sampled noise from either uniform or normal distributions to all parameters within $\bm{\sigma}$, the initial configuration of the robot in each episode, $\bm{q}_{\text{init}}$, the mass of the robot's base, $m_{B}$, teh fiction coefficient between the robot's foot and the ground, $\mu_{\text{fric}}$, $K_{p}$ and $K_{p}$. The details of this is presented in Supplementary Table 4. Additionally, to ensure that a rich variation of $\bm{U}^{\text{cmd}}$ is experienced during training, randomly sampled gaits, velocity commands and velocity change durations (to achieve random acceleration), $t_{\text{acc}}$, are implemented during training. These are within defined maximum, $t_{\text{acc}}^{\text{max}}=0.5\text{s}$, and minimum, $t_{\text{acc}}^{\text{min}}=0\text{s}$, acceleration durations. Further sampling details are also presented in Supplementary Table 4. 
\begin{table}[h]
\centering
\caption{Noise and Sampling Distributions.} \label{tab:dist}
\vspace{-1mm}
\begin{tabular}{lc}
\toprule
\multicolumn{1}{l}{\parbox{1.2cm}{\centering Parameter}} & \multicolumn{1}{c}{\parbox{1.2cm}{\centering Distribution}} \\ \vspace{-2mm} \\
\midrule 
$\bm{\omega}_{B}$, $\dot{\bm{v}}_{B}$ & $0.015\mathcal{N}(0,1)$  \\
$\bm{q}$ & $0.005\mathcal{N}(0,1)$ \\
$\dot{\bm{q}}$ & $0.15\mathcal{N}(0,1)$ \\
$\bm{\tau}$, $\bm{f}_{\text{frc}}$ & $\mathcal{N}(0,1)$ \\
$\mu_{\text{fric}}$ & $\max(0.4 ,\min(0.6+0.5\mathcal{N}(0,1), 1))$ \\
$m_{B}$ & $\max(-1 ,\min(\mathcal{N}(0,1), 3))$\\
$K_{p}$ & $K_{p}\max(0.9 ,\min(1+0.05\mathcal{N}(0,1), 1.1))$\\
$K_{d}$ & $K_{d}\max(0.9 ,\min(1+0.05\mathcal{N}(0,1), 1.1))$\\
$v_{x}^{\text{cmd}}$ & $\mathcal{U}(0,1.5)$ \\
$\omega_{z}^{\text{cmd}}$ & $\mathcal{U}(-1,1)$ \\
$\Gamma$ & $\mathcal{U}(0,6)$ \\
$t_{\text{acc}}$ & $\mathcal{U}(t_{\text{acc}}^{\text{min}},t_{\text{acc}}^{\text{max}})$ \\
\bottomrule
\end{tabular}
\vspace{-2mm}
\end{table}

\section{Supplementary Video 1}
\label{sup:video1}
Link: \url{https://youtu.be/NwHoB7pErYQ}

\section{Supplementary Video 2}
\label{sup:video2}
Link: \url{https://youtu.be/-DfkDFA3KkI}

\section{Supplementary Video 3}
\label{sup:video3}
Link: \url{https://youtu.be/y4KnzMEdf78}

\section{Supplementary Video 4}
\label{sup:video4}
Link: \url{https://youtu.be/I02DQ1RGdyw}

\section{Supplementary Video 5}
\label{sup:video5}
Link: \url{https://youtu.be/f6CqJ7gb3ZM}

\end{document}